%% file: main.tex
\documentclass{article}

\usepackage[main,final,nonatbib]{neurips_2026}
\usepackage{amsmath}
\usepackage{float}
\usepackage[numbers,compress]{natbib}
\usepackage[utf8]{inputenc} 
\usepackage[T1]{fontenc}    
\usepackage{url}            
\usepackage{booktabs}       
\usepackage{amsfonts}       
\usepackage{nicefrac}       
\usepackage{microtype}      
\usepackage{xcolor}         
\usepackage[table]{xcolor}
\usepackage[colorlinks=true,citecolor=citeblue]{hyperref}
\definecolor{lightred}{RGB}{255, 200, 200}
\definecolor{lightorange}{RGB}{255, 230, 180}
\definecolor{citeblue}{RGB}{87,145,220}
\usepackage{booktabs}
\usepackage{multirow}
\usepackage{graphicx}
\usepackage{pifont}
\usepackage{wrapfig}
\usepackage{longtable}

\title{ACD-GS: Asymmetric Curvature-aware Densification for 3D Gaussian Splatting}

\author{%
  Houqiang Zhong$^{1}$\thanks{Equal contributions.}\quad
  Tianchi Zhu$^{1,3}$\footnotemark[1]\quad
  Qiang Hu$^{2}$\thanks{Corresponding authors.} \quad 
  Xiaoyun Zhang$^{2}$ \\
  \textbf{Li Song}$^{1}$\footnotemark[2]\quad
  \textbf{Wenjun Zhang}$^{1}$ \\
  $^{1}$ School of Information Science and Electronic Engineering \\
  $^{2}$ Cooperative Medianet Innovation Center \\
  $^{3}$ SJTU Paris Elite Institute of Technology \\
  Shanghai Jiao Tong University, Shanghai, China \\
  \texttt{\{zhonghouqiang,zhutianchi,qiang.hu,xiaoyun.zhang\}@sjtu.edu.cn} \\
  \texttt{\{song\_li,zhangwenjun\}@sjtu.edu.cn}
}

\begin{document}

\maketitle

\begin{abstract}
3D Gaussian Splatting (3DGS) has emerged as a powerful paradigm for real-time photorealistic novel view synthesis. However, its training-time densification remains largely heuristic, often resulting in redundant Gaussian proliferation and inaccurate geometric reconstruction. Existing methods predominantly rely on first-order screen-space gradients to trigger cloning and splitting, which fail to distinguish true geometric discontinuities from high-frequency textures or noisy residuals. To address this limitation, we propose \textbf{ACD-GS}, an Asymmetric Curvature-aware Densification framework for compact and geometry-aware 3D Gaussian optimization. Our key insight is that second-order depth responses provide a more reliable structural prior for density control, enabling asymmetric regulation of cloning and splitting to better capture geometric discontinuities. Building on this, we introduce a multi-view photometric consistency constraint to suppress transient single-view errors, together with a capacity-aware adaptive regulation strategy that balances densification and pruning during training. Unlike post-hoc compression or quantization-based approaches, ACD-GS directly prevents redundant Gaussian generation during training, yielding inherently compact representations. Extensive experiments demonstrate that our method reduces nearly \textbf{60\%} of Gaussian count and storage while maintaining competitive rendering fidelity, consistently outperforming existing methods across multiple datasets. Our code is available in 
\url{https://waveviewer.github.io/ACD-GS/}.

\end{abstract}

\input{01intro_v8}

\input{02relatedworks_v8}
\input{03method_v8}
\input{04Experiments_v8}

\input{05Conclusion_v8}
\label{page:main_end}

\clearpage
\begin{ack}
This work is supported by National Natural Science Foundation of China(62431015, 62571322, 62571317),STCSM ( 24ZR1432000, 22DZ2229005), 111 plan (BP0719010),  and State Key Laboratory of UHD Video and Audio Production and Presentation.
\end{ack}
{
\bibliographystyle{IEEEtran}
\small
\bibliography{reference_v8}
}

\newpage
\renewcommand{\theHtable}{appendix.\arabic{table}}
\input{appendix_v8}

\clearpage
\input{checklist.tex}
\end{document}

%% file: 01intro_v8.tex
\section{Introduction}

3D Gaussian Splatting (3DGS)~\cite{3dgs} has rapidly emerged as a powerful paradigm for real-time photorealistic novel view synthesis. Unlike implicit neural radiance fields, 3DGS represents a scene as an explicit collection of anisotropic Gaussian primitives and renders them through differentiable rasterization, achieving both high rendering fidelity and remarkable efficiency. Benefiting from its explicit representation and fast optimization, 3DGS has inspired extensive advances in anti-aliasing~\cite{mipsplatting,prismgs,AnalyticSplatting}, surface reconstruction~\cite{aligngs,sugar,2dgs,gsmesh,pgsr}, dynamic scene modeling~\cite{4DGC,4dgcpro,spacetimegs,4dgs,luiten2023dynamic}, compression~\cite{compact_3dgs,3DGSzip,hac,contextgs,hac++}, and hybrid implicit-explicit representations~\cite{scaffoldgs,octreegs,gsembedding,splatnet,gnf,tang2026neuralgs}. Despite these advances, the efficiency and scalability of 3DGS still critically depend on how Gaussian primitives are created and evolved during optimization.
\begin{figure}[t]
    \centering
    \includegraphics[width=\linewidth]{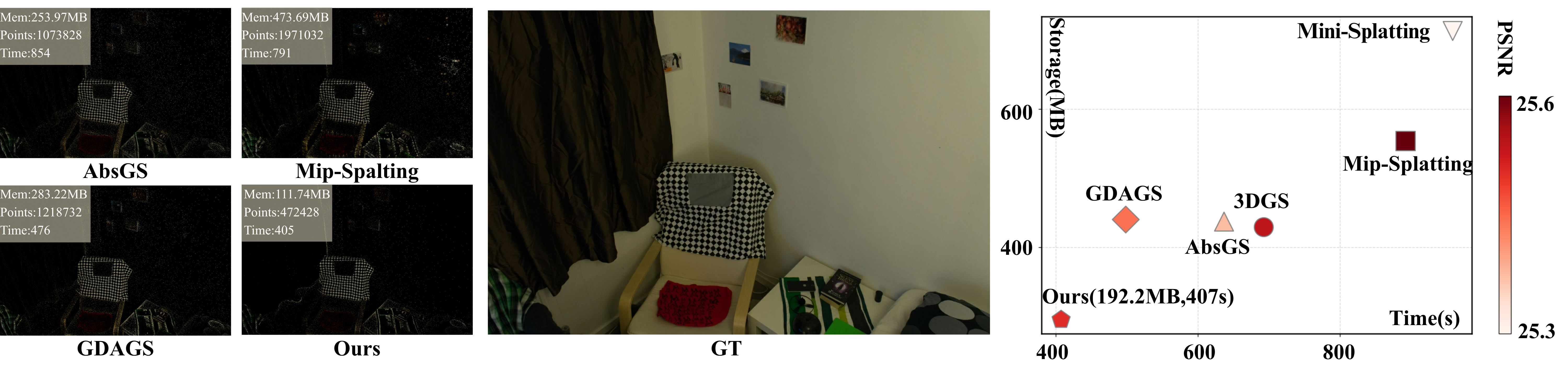}
    \caption{
    The point-cloud visualizations show our \textbf{ACD-GS} produces sparser Gaussians while preserving scene structure. The right plot (training time vs. storage, color-coded by PSNR) confirms that we achieve lower storage and faster training without sacrificing rendering quality.
    }
    \setlength{\belowcaptionskip}{2pt}
    \label{fig:teaser}
\end{figure}

In standard 3DGS, densification is mainly driven by accumulated screen-space positional gradients, where Gaussians are cloned or split once the gradient magnitude exceeds predefined thresholds. While simple and effective in practice, this mechanism remains largely geometry-agnostic, as first-order gradients primarily reflect local optimization pressure rather than underlying scene structure. Consequently, regions with high-frequency textures, smoothly varying surfaces, and sharp geometric boundaries may all produce similarly large gradients despite requiring fundamentally different density-control behaviors. A flat textured wall, for example, should avoid excessive Gaussian proliferation, whereas thin structures and depth discontinuities require localized geometric refinement. This reveals an intrinsic ambiguity in gradient-driven densification, where first-order photometric signals alone are insufficient for reliable density control.

This ambiguity leads to a persistent trade-off in existing densification strategies~\cite{iadc,pixel_gs,absgs,gdags,steep_gs}. Aggressive thresholds tend to trigger excessive Gaussian growth in textured regions, resulting in redundant primitives, increased storage overhead, unstable optimization, and floating artifacts near foreground-background transitions. Conversely, conservative thresholds often fail to reconstruct fine structures and sharp depth discontinuities. Recent approaches attempt to improve densification stability by incorporating pixel-aware gradients~\cite{pixel_gs}, absolute gradient magnitudes~\cite{absgs}, or gradient-direction statistics~\cite{gdags,steep_gs}. Although these methods alleviate optimization instability to some extent, they still rely primarily on first-order, single-view photometric signals and therefore remain insufficient for disentangling true geometric complexity from texture-induced residuals. As a result, existing methods struggle to simultaneously achieve compact representations, geometric fidelity, and stable optimization. These limitations motivate the need for a geometry-aware density-control prior that directly reflects scene structure rather than relying solely on photometric optimization cues.

In this work, we propose \textbf{ACD-GS}, an Asymmetric Curvature-aware Densification framework that reformulates Gaussian densification as a geometry-aware optimization process. Our key insight is that second-order depth responses provide a more reliable structural cue for identifying geometric discontinuities than first-order screen-space gradients. Compared with gradient magnitude alone, depth curvature is more sensitive to local geometric changes such as occlusion boundaries, thin structures, and high-curvature regions, while remaining less responsive to purely planar textured areas. Based on this observation, ACD-GS introduces an asymmetric densification mechanism that explicitly decouples cloning and splitting behaviors according to curvature-aware structural priors. Specifically, cloning is suppressed near depth discontinuities to avoid ambiguous foreground-background duplication and floating artifacts, whereas splitting is promoted to refine boundary-crossing Gaussians into finer geometric primitives.

To further resolve the ambiguity between geometry and appearance, we introduce a multi-view photometric error gating strategy that enforces cross-view consistency instead of relying on volatile single-view residuals. Furthermore, we design a capacity-constrained adaptive optimization scheme that dynamically regulates primitive growth and lifecycle-aware pruning throughout training. Together, these designs establish a unified geometry-aware density-control framework that enables native sparsification, where redundant Gaussian primitives are prevented during training rather than removed through post-hoc compression or quantization pipelines. Extensive experiments demonstrate that ACD-GS reduces nearly \textbf{60\%} of Gaussian primitives and storage while maintaining competitive rendering fidelity across multiple datasets, consistently achieving a favorable quality and compactness trade-off compared with existing methods.


Our contributions are summarized as follows:
\begin{itemize}
    \item We propose \textbf{ACD-GS}, an asymmetric curvature-aware densification framework that leverages second-order depth priors to explicitly decouple cloning and splitting behaviors for geometry-aware Gaussian evolution.
      \item We introduce a multi-view photometric error gating strategy and a capacity-constrained adaptive optimization scheme to suppress transient photometric noise, regulate primitive growth, and encourage compact yet high-fidelity representations.
       \item Extensive experiments demonstrate that ACD-GS significantly reduces Gaussian count and model storage natively during training while maintaining competitive rendering quality across multiple datasets.
\end{itemize}

%% file: 02relatedworks_v8.tex
\section{Related Work}
\label{sec:related}

\subsection{3D Gaussian Splatting Representation}
3D Gaussian Splatting (3DGS)~\cite{3dgs} represents a scene as anisotropic Gaussian primitives and renders them through differentiable rasterization. Its explicit formulation has inspired extensions for anti-aliasing~\cite{mipsplatting,prismgs,AnalyticSplatting}, surface reconstruction~\cite{aligngs,sugar,2dgs,gsmesh,pgsr}, dynamic scenes~\cite{4DGC,4dgcpro,spacetimegs,4dgs,luiten2023dynamic}, and appearance modeling~\cite{HyperGaussians,hypergs,thermalgaussian,7dgs}. Hybrid representations such as Scaffold-GS~\cite{scaffoldgs}, SplatNet~\cite{splatnet}, and Octree-GS~\cite{octreegs} further organize primitives with learned features or hierarchical structures, while implicit fields~\cite{nerf,instantngp,mobilenerf,snerg} provide complementary scene parameterizations. These methods mainly change representation or rendering; the number and placement of Gaussians still depend on training-time density control, which is the focus of this work. Because all of them retain a primitive-based rendering pipeline, their efficiency and fidelity are also affected by how many Gaussians remain after optimization.

\subsection{Gradient-Driven Density Control}
Standard 3DGS~\cite{3dgs} densifies Gaussians using accumulated screen-space positional gradients. Pixel-GS~\cite{pixel_gs} accounts for pixel coverage, AbsGS~\cite{absgs} uses absolute gradients to reduce cancellation, and GDAGS~\cite{gdags} incorporates gradient-direction information; other methods add external constraints~\cite{iadc,steep_gs}. FastGS~\cite{fastgs} instead estimates Gaussian importance from multi-view reconstruction errors inside projected footprints and uses this score for both densification and pruning. ACD-GS shares the use of multi-view evidence but assigns it a different role: rendered-depth curvature modulates cloning and splitting, while center-projected RGB residuals only verify whether a candidate is consistently under-reconstructed across views. This separates structural guidance from photometric verification and avoids using a footprint error score as the direct densification criterion.

\subsection{3DGS Compression}
Post-hoc compression reduces Gaussian storage through significance-based pruning and vector quantization~\cite{lightgaussian,eagles,compact_3dgs,trimming_fat,OccluGaussian,PUP3D-GS}, while context models and entropy coding target rate--distortion efficiency~\cite{hac,contextgs,rdogaussian,pcgs,cat3dgs}. Mini-Splatting~\cite{mini_splatting} studies compact representations under constrained primitive budgets. These approaches mainly operate after or around training and may require codebooks or fine-tuning. ACD-GS instead reduces redundant primitive generation during optimization, without a post-hoc quantization stage. Because it changes where capacity is allocated rather than how parameters are encoded, it can be combined with downstream compression; this combination is not evaluated here. This distinction also delimits the scope of our experiments, which measure training-time sparsification rather than gains from entropy coding or quantization.

%% file: 03method_v8.tex
\section{Method}
\label{sec:method}

\subsection{Problem Formulation and Overview}
\label{sec:problem_overview}

\begin{figure}[t]
    \centering
    \includegraphics[width=1\linewidth]{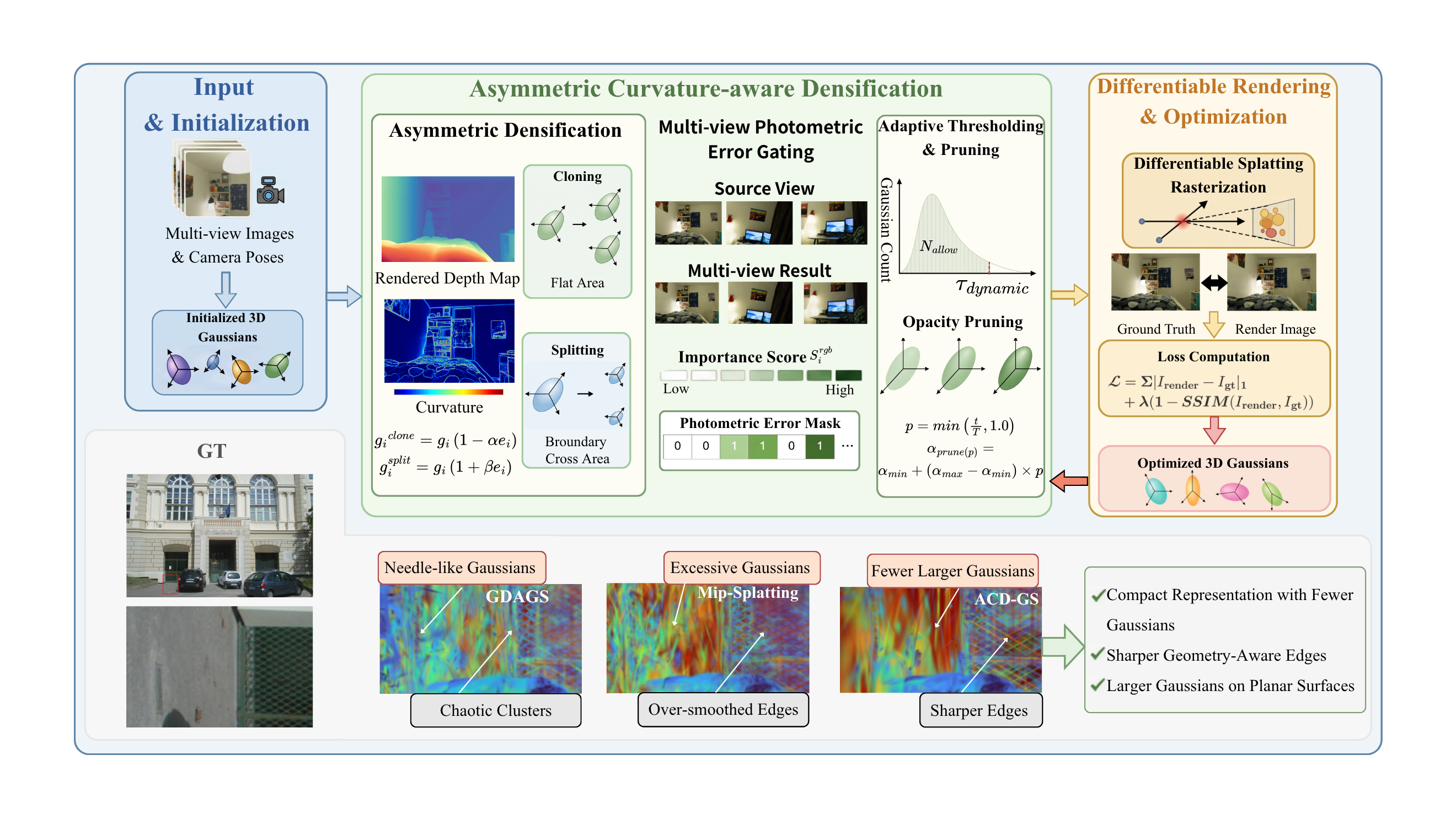}
    \caption{
    Overview of ACD-GS. Curvature-guided asymmetric densification suppresses cloning and promotes splitting near depth boundaries; multi-view photometric gating verifies candidates; capacity-aware thresholding and opacity pruning regulate growth. Depth is rendered from the current Gaussians rather than input.
    }
    \label{fig:overview}
\end{figure}

3D Gaussian Splatting (3DGS) represents a scene as a set of anisotropic Gaussian primitives
$\mathcal{G}=\{G_i\}_{i=1}^{N}$ optimized through differentiable alpha blending.
During training, standard 3DGS densifies visible Gaussians according to their accumulated
view-space positional gradients:
\begin{equation}
    g_i =
    \sum
    \left\|
    \frac{\partial \mathcal{L}}
    {\partial \boldsymbol{\mu}^{2D}_i}
    \right\|,
\end{equation}
where $\boldsymbol{\mu}^{2D}_i$ is the projected 2D center of $G_i$.
A Gaussian is cloned or split once $g_i$ exceeds a fixed threshold
$\tau_{\mathrm{grad}}$, with the operation determined by scale-based heuristics. However, such first-order gradient-driven densification is geometry-agnostic.
Large photometric gradients may be caused by flat textures, smooth slopes, or true depth
discontinuities, although these cases require different density-control behaviors.
This ambiguity often leads to redundant primitives on planar or textured regions, while sharp
boundaries and thin structures may remain under-refined.



We propose \textbf{Asymmetric Curvature-aware Densification for 3D Gaussian Splatting} (\textbf{ACD-GS}), a plug-and-play density-control module that replaces the heuristic densification step in 3DGS to prevent redundant primitive generation at its source (Fig.~\ref{fig:overview}). At its core, ACD-GS utilizes a second-order depth curvature prior, which is highly responsive to structural discontinuities but robust to planar textures, to asymmetrically suppress cloning near geometric edges while actively promoting the splitting of boundary-crossing Gaussians. To further resolve the geometry-texture trade-off, a multi-view photometric gate restricts growth to candidates exhibiting consistent cross-view RGB residuals, effectively filtering transient noise. Finally, capacity-constrained adaptive thresholding and lifecycle-aware opacity pruning dynamically regulate the global primitive scale under a scene-aware quota. By fundamentally curbing proliferation during optimization, ACD-GS yields highly compact representations without the need to compress an already overgrown Gaussian set post-training.

\subsection{Curvature-Guided Asymmetric Densification}
\label{sec:curvature_densification}

To distinguish geometric structures from texture-induced photometric variations, we introduce a
curvature-guided structural prior.
Given a rendered depth map $D$, we compute a curvature proxy using the magnitude of the discrete
Laplacian:
\begin{equation}
    D^{(2)} = \left| \nabla^2 D \right|.
\end{equation}
This second-order response is less sensitive to high-frequency textures on locally planar surfaces,
while producing strong responses at depth discontinuities and high-curvature regions.

To improve robustness against isolated depth spikes, we clip and normalize the curvature response:
\begin{equation}
    \tilde{D}^{(2)} =
    \operatorname{clip}
    \left(
    D^{(2)}, 0,
    \mu_{D^{(2)}} + 3\sigma_{D^{(2)}}
    \right),
\end{equation}
\begin{equation}
    \hat{D}^{(2)} =
    \frac{\tilde{D}^{(2)}}{\max(\tilde{D}^{(2)})+\epsilon}.
\end{equation}
Weak responses are further suppressed by soft thresholding:
\begin{equation}
    E =
    \frac{
    \max(\hat{D}^{(2)}-\tau,0)
    }{
    1-\tau
    },
\end{equation}
yielding a depth-edge confidence map $E\in[0,1]$. We use $\tau=0.15$.
For each visible Gaussian, we sample this confidence at its projected center:
\begin{equation}
    e_i = E(\boldsymbol{\mu}_{i}^{2D}), \quad e_i \in [0,1].
\end{equation}

Standard 3DGS uses the same gradient statistic for both cloning and splitting, despite their
different geometric effects.
Cloning increases local representation capacity and is useful for texture-rich smooth regions, but may
duplicate ambiguous primitives near foreground-background transitions.
Splitting, instead, decomposes large Gaussians and is more suitable for refining primitives that span
multiple depth layers.

We therefore introduce operation-specific asymmetric gradient modulation:
\begin{equation}
    g_i^{\mathrm{clone}} = g_i(1-\alpha e_i),
    \quad
    g_i^{\mathrm{split}} = g_i(1+\beta e_i),
\end{equation}
where $\alpha=0.8$ and $\beta=1.5$ control the suppression and enhancement strengths.
Near structural boundaries, cloning is suppressed to avoid duplicating ambiguous primitives, while
splitting is amplified to encourage fine-grained boundary refinement.
In smooth regions, the modulated gradients reduce to the original $g_i$, preserving the baseline
capacity for appearance modeling.

\subsection{Multi-view Photometric Error Gating}
\label{sec:multiview_photometric_gating}
 Although curvature provides a useful structural prior, photometric signals remain necessary for
densifying texture-rich flat regions.
However, single-view residuals may be unstable due to occlusion, view-dependent effects, or transient
optimization noise.
We therefore verify photometric necessity across multiple views.

Before each densification step, we sample $K=8$ auxiliary cameras from the training set.
For each view $k$, we render an image $I_{\mathrm{render}}^{(k)}$ and compare it with the ground-truth
image $I_{\mathrm{gt}}^{(k)}$ using the mean absolute RGB residual:
\begin{equation}
    E_{\mathrm{rgb}}^{(k)}
    =
    \frac{1}{3}
    \sum_{c=1}^{3}
    \left|
    I_{\mathrm{render},c}^{(k)}
    -
    I_{\mathrm{gt},c}^{(k)}
    \right|
    \odot
    M_{\mathrm{valid}}^{(k)},
\end{equation}
where $M_{\mathrm{valid}}^{(k)}$ is the valid image mask.
The residual map is binarized as
\begin{equation}
    \mathcal{B}_{\mathrm{rgb}}^{(k)}
    =
    \mathbb{I}
    \left(
    E_{\mathrm{rgb}}^{(k)} > \tau_{\mathrm{rgb}}
    \right)
    \odot
    M_{\mathrm{valid}}^{(k)} .
\end{equation}

For each Gaussian $G_i$, we project its 3D center $\boldsymbol{\mu}_i$ to view $k$ and sample the
binary residual map:
\begin{equation}
    e_{i,k}^{\mathrm{rgb}}
    =
    \mathcal{B}_{\mathrm{rgb}}^{(k)}
    \left(
    \mathbf{p}_{i,k}
    \right),
\end{equation}
where $\mathbf{p}_{i,k}$ is the projected screen-space coordinate.
Let $v_{i,k}\in\{0,1\}$ indicate whether the projection lies in the valid image region.
The multi-view RGB error score is then
\begin{equation}
    S_{i}^{\mathrm{rgb}}
    =
    \frac{
    \sum_{k=1}^{K}
    v_{i,k} e_{i,k}^{\mathrm{rgb}}
    }{
    \max
    \left(
    \sum_{k=1}^{K}
    v_{i,k}, 1
    \right)
    }.
\end{equation}
We define the photometric gate as
\begin{equation}
    \mathcal{M}_{i}^{\mathrm{rgb}}
    =
    \mathbb{I}
    \left(
    S_{i}^{\mathrm{rgb}}
    \ge
    \tau_{\mathrm{mv}}
    \right).
\end{equation}
Only Gaussians satisfying $\mathcal{M}_{i}^{\mathrm{rgb}}=1$ are admitted into the densification
candidate pool.
This cross-view verification suppresses transient single-view residuals while preserving
under-reconstructed appearance details.
The valid-projection indicator checks image support and should not be interpreted as an explicit geometric occlusion test.

\subsection{Capacity-Constrained Adaptive Optimization}
\label{sec:adaptive_optimization}

Fixed densification and pruning thresholds are suboptimal because gradient distributions and
primitive scales vary across scenes and training stages.
We therefore regulate both primitive growth and removal according to the current optimization state.

\paragraph{Capacity-Adaptive Densification.}
At each densification interval, we estimate an adaptive threshold from the valid gradient distribution.
Let $N_{\mathrm{total}}$ denote the current number of Gaussians, and let
$\mathcal{G}_{\mathrm{valid}}=\{g_i \mid g_i>0\}$ contains the positive accumulated position gradients used for threshold estimation.
We impose a scene-dependent growth quota:
\begin{equation}
    N_{\mathrm{allow}}
    =
    \left\lfloor
    \mu N_{\mathrm{total}}
    \right\rfloor ,
\end{equation}
where $\mu=0.008$ (0.8\%) is the default quota ratio.
Let $M = |\mathcal{G}_{\mathrm{valid}}|$, and sort the candidate gradients as
$g_{(1)} \ge g_{(2)} \ge \cdots \ge g_{(M)}$.
The base adaptive threshold is
\begin{equation}
    \hat{\tau}_{\mathrm{dyn}}
    =
    \begin{cases}
        g_{(N_{\mathrm{allow}})}, & M > N_{\mathrm{allow}}, \\
        g_{(k_{\mathrm{top}})}, & M \le N_{\mathrm{allow}},
    \end{cases}
    \quad
    k_{\mathrm{top}}
    =
    \max(1,\lceil 0.1M\rceil).
\end{equation}
When the number of candidates exceeds the quota, this threshold directly bounds the number of
densification operations.
Otherwise, the top-$10\%$ fallback avoids selecting marginal gradients.
The final threshold is lower-bounded by a relaxed baseline threshold:
\begin{equation}
    \tau_{\mathrm{dyn}}
    =
    \max
    \left(
    0.8\tau_{\mathrm{base}},
    \hat{\tau}_{\mathrm{dyn}}
    \right).
\end{equation}

\paragraph{Lifecycle-Aware Opacity Pruning.}
We further adapt opacity pruning according to training progress.
Let
\begin{equation}
    p =
    \min
    \left(
    \frac{t}{T},
    1
    \right),
\end{equation}
where $t$ and $T$ are the current and total training iterations.
The opacity pruning threshold is annealed as
\begin{equation}
    \alpha_{\mathrm{prune}}(p)
    =
    \alpha_{\mathrm{min}}
    +
    (\alpha_{\mathrm{max}}-\alpha_{\mathrm{min}})p .
\end{equation}
This preserves exploratory low-opacity primitives in early training, while enforcing stricter removal
of redundant primitives near convergence.

Overall, ACD-GS couples structural densification, cross-view photometric verification, and
capacity-aware optimization.
The resulting training process produces compact Gaussian representations without relying on
post-hoc compression.

%% file: 04Experiments_v8.tex
\section{Experiments}

\vspace{0.51em}
\input{tables/deepblending_v8}

\textbf{Experimental Setup}. We evaluate ACD-GS on the Mip-NeRF 360 \cite{mipinerf360}, Tanks \& Temples \cite{tank}, and Deep Blending \cite{DeepBlending} datasets against vanilla 3DGS \cite{3dgs} and representative baselines including Mip-Splatting \cite{mipsplatting}, Mini-Splatting \cite{mini_splatting}, AbsGS \cite{absgs}, GDAGS \cite{gdags}, and Pixel-GS \cite{pixel_gs}. All models are trained for 30K iterations on a single NVIDIA RTX 5090 GPU. Our curvature-aware density control operates every 100 steps from iteration 600 until training completion to strictly regulate primitive proliferation. We report rendering metrics (PSNR, SSIM, LPIPS) and efficiency statistics (storage size, Gaussian count, training time). Further implementation details are provided in the Appendix.




\subsection{Quantitative Comparison}
\textbf{Rendering Fidelity.} 
On the Deep Blending dataset (Tab.~\ref{tab:Deepblending_combined}), ACD-GS achieves the best overall rendering quality among all evaluated methods, obtaining the highest PSNR (25.12) and SSIM (0.834). This demonstrates that our compact representation does not compromise the reconstruction of visually important structures. Compared with vanilla 3DGS, ACD-GS consistently produces sharper reconstructions while avoiding representation size bloat caused by unconstrained densification. For the large-scale indoor and outdoor scenes in Mip-NeRF 360 and Tanks \& Temples (Tab.~\ref{tab:tandt_mipnerf}), ACD-GS maintains highly competitive visual quality and matches or exceeds strong gradient-based baselines such as AbsGS and GDAGS. These results suggest that the proposed curvature-aware prior effectively preserves essential structural and textural details, enabling compactness without sacrificing rendering fidelity.

\textbf{Model Compactness and Training Efficiency.} 
Tab.~\ref{tab:effiency} reports the efficiency comparison in terms of storage footprint, Gaussian count, and training time. Across all three dataset groups, ACD-GS consistently achieves the smallest model size and the fastest training among the compared methods. This indicates that our native sparsification strategy controls redundancy during optimization rather than relying on post-hoc compression. Compared with GDAGS, ACD-GS reduces average storage from 360.6 MB to 151.9 MB on Deep Blending, corresponding to a 57.8\% reduction; from 707.1 MB to 274.3 MB on Mip-NeRF 360, corresponding to a 61.2\% reduction; and from 319.5 MB to 150.4 MB on Tanks \& Temples, corresponding to a 52.9\% reduction. The Gaussian count is reduced by roughly 60\% across the datasets, which further lowers memory usage and computational cost during training.

\textbf{Trade-off Advantage.} 
The key advantage of ACD-GS lies in its favorable quality and compactness trade-off. Recent density-control methods, such as Pixel-GS, often improve rendering quality by introducing excessive numbers of primitives, reaching 5.56M Gaussians on Mip-NeRF 360. In contrast, ACD-GS suppresses redundant Gaussian generation at its optimization source through curvature-aware asymmetric clone and split modulation and multi-view gating. As confirmed by the storage--quality plot in Fig.~\ref{fig:teaser}, ACD-GS occupies the most favorable region of the trade-off space, achieving competitive fidelity with substantially lower storage. This demonstrates that our method effectively separates geometrically necessary densification from texture-induced or view-specific noise, leading to a compact yet expressive Gaussian representation.

\input{tables/tandt_mipnerf_v8}
\input{tables/efficiency_v8}

\subsection{Qualitative Comparison}
\begin{figure}[t]
    \centering
    \includegraphics[width=\linewidth]{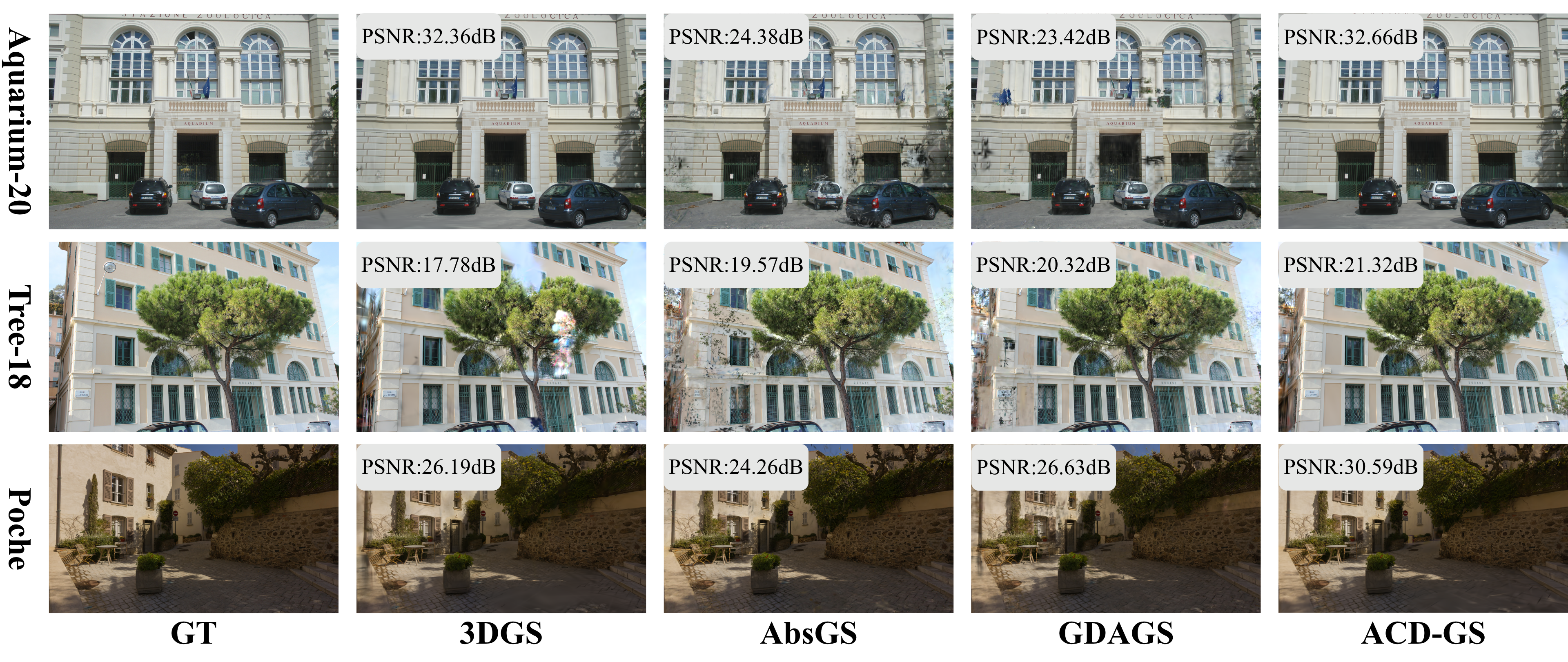}
    \caption{Qualitative comparison of novel view synthesis on the Deep Blending dataset.}
    \label{fig:rgb_comp}
\end{figure}

\textbf{Visual Rendering Fidelity.} 
As shown in Fig.~\ref{fig:rgb_comp}, ACD-GS achieves novel view synthesis quality that is highly comparable to, and occasionally surpasses, the unconstrained vanilla 3DGS model. Gradient-based baselines like AbsGS and GDAGS frequently suffer from texture degradation to maintain a lower point count, our method consistently preserves high-frequency details. Crucially, vanilla 3DGS is notoriously prone to generating severe floating artifacts in empty spaces (visibly evident in the sky region of the Tree scene) due to its geometric blindness. In contrast, by rigorously suppressing redundant densification at its source, ACD-GS effectively eliminates these floaters and reconstructs remarkably clean backgrounds and sharp object boundaries.

\begin{figure}[t]
    \centering
    \includegraphics[width=\linewidth]{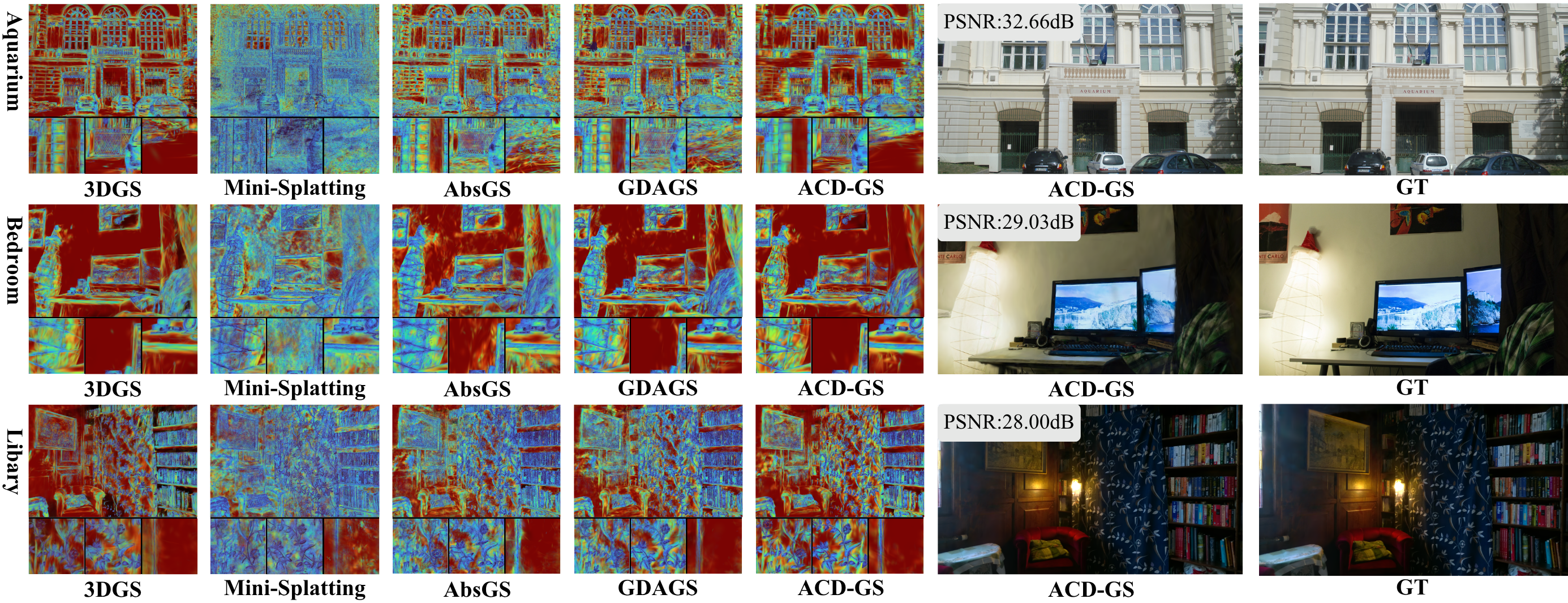}
    \caption{Visualization of Gaussians distribution}
    \label{fig:density_comp}
\end{figure}

\textbf{Primitive Distribution Analysis.} 
To uncover the structural rationale behind our efficiency and rendering quality, we visualize the spatial density of Gaussian primitives in Fig.~\ref{fig:density_comp}. Vanilla 3DGS exhibits extreme redundancy, indiscriminately populating flat homogeneous surfaces with overly dense primitives. Other density-control baselines often fail to cleanly separate textures from geometry, leading to either noisy distributions or over-pruned structures. Conversely, the primitive allocation in ACD-GS strongly aligns with the underlying scene geometry. 
Guided by our second-order curvature prior, ACD-GS allocates dense and highly precise Gaussians strictly along structural edges and complex details. Meanwhile, it elegantly covers planar regions (such as the building facades in the Aquarium scene and the walls in the Bedroom scene) with much larger and highly sparse primitives. This structurally aware evolution precisely explains why ACD-GS achieves extreme model compactness and fewer floating artifacts without compromising visual fidelity. More results are shown in Appendix.

\subsection{Parameter Sensitivity Analysis}
Tab.~\ref{tab:parametre10} analyzes the sensitivity of ACD-GS to $\alpha$ and $\beta$. We vary one parameter at a time: $\alpha$ with $\beta=1.5$ fixed, and $\beta$ with $\alpha=0.8$ fixed. Increasing $\alpha$ from 0.5 to 0.8 improves PSNR and SSIM while reducing LPIPS and storage. Setting $\alpha=1.0$ yields the highest PSNR and SSIM, but also a larger LPIPS and model size. For $\beta$, the default $\beta=1.5$ gives the best PSNR, SSIM, LPIPS, and storage among the tested values. Overall, $(\alpha,\beta)=(0.8,1.5)$ provides the best quality--compactness balance, while $\alpha=1.0$ offers higher PSNR and SSIM at a larger model size.
\input{tables/alpha_beta_ablation_v8}

\vspace{-0.5em}
\subsection{Ablation Studies}
\label{sec:ablation}
We conduct an incremental ablation study integrating our three core modules: Curvature-Guided Asymmetric Densification (CAD), Multi-view Photometric Error Gating (MPG), and Capacity-Constrained Adaptive Optimization (CAO). We evaluate the model storage requirements and rendering fidelity on the DeepBlending dataset.  As shown in Tab.~\ref{tab:ablation}, the baseline 3DGS suffers from severe memory bloat (360.6 MB) due to unconstrained primitive proliferation. Integrating the CAD module effectively addresses this by penalizing redundant cloning in flat regions and promoting boundary-aware splitting, which immediately reduces storage to 302.4 MB while improving PSNR. Subsequently, incorporating MPG further slashes the memory footprint to 159.7 MB. By demanding cross-view residual consensus, MPG prevents the model from blindly overfitting to transient single-view photometric noise, concurrently yielding strict improvements in SSIM and LPIPS. Through dynamic growth quotas and lifecycle-aware pruning, CAO curtails floating artifacts and reduces the final model size to a remarkable 151.9 MB, representing a 58\% reduction from the baseline.

\input{sensitivity_v8}

%% file: tables/deepblending_v8.tex
\begin{table}[t]
\centering
\caption{Quantitative Comparison on Deep Blending}
\label{tab:Deepblending_combined}
\small
\setlength{\tabcolsep}{2.5pt}
\begin{tabular}{@{}lcccccccc@{}}
\toprule
& \multicolumn{4}{c}{\textbf{Aquarium}} & \multicolumn{4}{c}{\textbf{Bedroom}} \\
\cmidrule(lr){2-5} \cmidrule(lr){6-9}
Method & SSIM$\uparrow$ & PSNR$\uparrow$ & LPIPS$\downarrow$ & Size(MB)$\downarrow$ & SSIM$\uparrow$ & PSNR$\uparrow$ & LPIPS$\downarrow$ & Size(MB)$\downarrow$ \\
\midrule
3DGS & \cellcolor{lightorange}0.869 & 22.66 & \cellcolor{lightorange}0.161 & 177.44 & \cellcolor{lightorange}0.907 & 28.86 & \cellcolor{lightorange}0.274 & 316.93\\
Mip-Splatting & 0.851 & 22.52 & 0.176 & 272.31 & 0.904 & 28.68 & 0.274 & 473.69\\
Mini-Splatting & 0.851 & 23.81 & 0.166 & 810.17 & 0.901 & 28.28 & 0.295 & 981.14\\
AbsGS & 0.806 & 21.82 & 0.229 & 183.16 & 0.901 & 28.85 & 0.275 & \cellcolor{lightorange}253.97\\
GDAGS & 0.842 & 22.18 & 0.186 & \cellcolor{lightorange}159.86 & 0.907 & \cellcolor{lightred}29.05 & \cellcolor{lightred}0.270 & 283.22\\
Pixel-GS & 0.865 & \cellcolor{lightorange}25.77 & \cellcolor{lightred}0.158 & 268.57 & 0.904 & 28.82 & 0.275 & 719.19\\
ACD-GS & \cellcolor{lightred}0.874 & \cellcolor{lightred}26.42 & 0.166 & \cellcolor{lightred}41.76 & \cellcolor{lightred}0.907 & \cellcolor{lightorange}29.04 & 0.290 & \cellcolor{lightred}111.74\\
\midrule
& \multicolumn{4}{c}{\textbf{CreepyAttic}} & \multicolumn{4}{c}{\textbf{Hugo}} \\
\cmidrule(lr){2-5} \cmidrule(lr){6-9}
Method & SSIM$\uparrow$ & PSNR$\uparrow$ & LPIPS$\downarrow$ & Size(MB)$\downarrow$ & SSIM$\uparrow$ & PSNR$\uparrow$ & LPIPS$\downarrow$ & Size(MB)$\downarrow$ \\
\midrule
3DGS & \cellcolor{lightred}0.892 & \cellcolor{lightorange}30.52 & 0.225 & 758.95 & \cellcolor{lightorange}0.843 & \cellcolor{lightorange}23.85 & \cellcolor{lightorange}0.139 & 321.95\\
Mip-Splatting & 0.886 & 30.16 & 0.221 & 1142.25 & 0.830 & 22.80 & 0.149 & 494.73\\
Mini-Splatting & 0.868 & 29.41 & 0.280 & 1172.18 & 0.841 & 23.24 & 0.143 & 873.89\\
AbsGS & 0.882 & 30.16 & \cellcolor{lightorange}0.220 & \cellcolor{lightorange}704.01 & 0.789 & 22.58 & 0.180 & 344.76\\
GDAGS & 0.889 & 30.43 & \cellcolor{lightred}0.214 & 711.70 & 0.821 & 23.39 & 0.160 & \cellcolor{lightorange}318.72\\
Pixel-GS & 0.884 & 30.24 & 0.222 & 1558.26 & 0.836 & 23.43 & \cellcolor{lightred}0.139 & 496.32\\
ACD-GS & \cellcolor{lightorange}0.891 & \cellcolor{lightred}30.70 & 0.244 & \cellcolor{lightred}165.98 & \cellcolor{lightred}0.847 & \cellcolor{lightred}24.45 & 0.161 & \cellcolor{lightred}110.88\\
\midrule
& \multicolumn{4}{c}{\textbf{Library}} & \multicolumn{4}{c}{\textbf{Museum-1}} \\
\cmidrule(lr){2-5} \cmidrule(lr){6-9}
Method & SSIM$\uparrow$ & PSNR$\uparrow$ & LPIPS$\downarrow$ & Size(MB)$\downarrow$ & SSIM$\uparrow$ & PSNR$\uparrow$ & LPIPS$\downarrow$ & Size(MB)$\downarrow$ \\
\midrule
3DGS & \cellcolor{lightred}0.781 & \cellcolor{lightorange}23.87 & \cellcolor{lightred}0.352 & \cellcolor{lightorange}501.54 & 0.836 & \cellcolor{lightorange}23.32 & \cellcolor{lightorange}0.135 & 567.26\\
Mip-Splatting & 0.763 & 23.32 & 0.358 & 829.34 & 0.836 & 23.28 & 0.135 & 807.78\\
Mini-Splatting & 0.762 & 23.33 & 0.394 & 1129.81 & \cellcolor{lightred}0.845 & 23.20 & 0.135 & 938.93\\
AbsGS & 0.731 & 23.49 & 0.416 & 738.66 & 0.793 & 21.73 & 0.169 & 568.56\\
GDAGS & 0.768 & 23.72 & \cellcolor{lightorange}0.357 & 568.79 & 0.834 & 22.69 & 0.138 & \cellcolor{lightorange}554.20\\
Pixel-GS & 0.777 & 23.73 & 0.358 & 998.51 & 0.835 & 23.15 & \cellcolor{lightred}0.134 & 841.11\\
ACD-GS & \cellcolor{lightorange}0.780 & \cellcolor{lightred}24.10 & 0.359 & \cellcolor{lightred}245.24 & \cellcolor{lightred}0.845 & \cellcolor{lightred}23.49 & 0.145 & \cellcolor{lightred}223.28\\
\midrule
& \multicolumn{4}{c}{\textbf{Museum-2}} & \multicolumn{4}{c}{\textbf{Tree}} \\
\cmidrule(lr){2-5} \cmidrule(lr){6-9}
Method & SSIM$\uparrow$ & PSNR$\uparrow$ & LPIPS$\downarrow$ & Size(MB)$\downarrow$ & SSIM$\uparrow$ & PSNR$\uparrow$ & LPIPS$\downarrow$ & Size(MB)$\downarrow$ \\
\midrule
3DGS & 0.897 & 25.34 & 0.099 & 333.72 & 0.797 & 20.75 & 0.170 & 223.22\\
Mip-Splatting & 0.896 & 25.81 & 0.098 & 477.92 & 0.789 & 21.06 & 0.176 & 289.15\\
Mini-Splatting & \cellcolor{lightred}0.905 & \cellcolor{lightorange}27.50 & \cellcolor{lightred}0.089 & 877.42 & \cellcolor{lightorange}0.804 & \cellcolor{lightorange}21.98 & 0.171 & 798.00\\
AbsGS & 0.880 & 25.48 & 0.111 & 310.55 & 0.746 & 21.03 & 0.210 & 216.07\\
GDAGS & 0.899 & 25.64 & 0.095 & \cellcolor{lightorange}299.60 & 0.770 & 21.26 & 0.205 & \cellcolor{lightorange}192.28\\
Pixel-GS & 0.898 & 26.58 & \cellcolor{lightorange}0.093 & 561.85 & 0.796 & 21.76 & \cellcolor{lightorange}0.167 & 327.29\\
ACD-GS & \cellcolor{lightorange}0.900 & \cellcolor{lightred}27.61 & 0.106 & \cellcolor{lightred}121.68 & \cellcolor{lightred}0.805 & \cellcolor{lightred}22.22 & \cellcolor{lightred}0.165 & \cellcolor{lightred}84.78\\
\midrule
& \multicolumn{4}{c}{\textbf{Yellowhouse}} & \multicolumn{4}{c}{\textbf{Mean}} \\
\cmidrule(lr){2-5} \cmidrule(lr){6-9}
Method & SSIM$\uparrow$ & PSNR$\uparrow$ & LPIPS$\downarrow$ & Size(MB)$\downarrow$ & SSIM$\uparrow$ & PSNR$\uparrow$ & LPIPS$\downarrow$ & Size(MB)$\downarrow$ \\
\midrule
3DGS & 0.655 & 17.11 & \cellcolor{lightorange}0.295 & \cellcolor{lightorange}176.08 & \cellcolor{lightorange}0.831 & 24.03 & \cellcolor{lightorange}0.205 & 375.23\\
Mip-Splatting & 0.641 & 16.75 & 0.316 & 238.22 & 0.822 & 23.82 & 0.211 & 558.38\\
Mini-Splatting & 0.639 & 16.27 & 0.316 & 749.61 & 0.824 & 24.11 & 0.221 & 925.68\\
AbsGS & 0.581 & 16.29 & 0.352 & 236.54 & 0.790 & 23.49 & 0.240 & 395.14\\
GDAGS & 0.637 & 16.63 & 0.308 & 180.94 & 0.819 & 23.89 & 0.215 & \cellcolor{lightorange}363.26\\
Pixel-GS & \cellcolor{lightred}0.667 & \cellcolor{lightorange}17.61 & \cellcolor{lightred}0.282 & 267.99 & 0.829 & \cellcolor{lightorange}24.56 & \cellcolor{lightred}0.203 & 671.01\\
ACD-GS & \cellcolor{lightorange}0.661 & \cellcolor{lightred}18.09 & 0.297 & \cellcolor{lightred}95.58 & \cellcolor{lightred}0.834 & \cellcolor{lightred}25.12 & 0.215 & \cellcolor{lightred}133.44\\
\bottomrule
\end{tabular}
\end{table}

%% file: tables/tandt_mipnerf_v8.tex
\begin{table}[t]
\centering
\caption{Quantitative Comparison on Tanks \& Temples and Mip-NeRF 360}
\label{tab:tandt_mipnerf}
\small
\setlength{\tabcolsep}{2.5pt}
\begin{tabular}{@{}lcccccccc@{}}
\toprule
& \multicolumn{4}{c}{\textbf{Train}} & \multicolumn{4}{c}{\textbf{Truck}} \\
\cmidrule(lr){2-5} \cmidrule(lr){6-9}
Method & SSIM$\uparrow$ & PSNR$\uparrow$ & LPIPS$\downarrow$ & Size(MB)$\downarrow$ & SSIM$\uparrow$ & PSNR$\uparrow$ & LPIPS$\downarrow$ & Size(MB)$\downarrow$ \\
\midrule
3DGS & 0.8202 & \cellcolor{lightorange}22.04 & 0.1963 & 258.35 & 0.8851 & 25.47 & 0.1418 & 487.82\\
Mip-Splatting & \cellcolor{lightorange}0.8266 & 21.87 & \cellcolor{lightorange}0.1893 & 354.45 & \cellcolor{lightred}0.8930 & \cellcolor{lightred}25.74 & \cellcolor{lightorange}0.1231 & 784.70\\
Mini-Splatting & 0.7099 & 18.77 & 0.3234 & 937.32 & 0.8573 & 23.66 & 0.1583 & 1084.63\\
AbsGS & 0.8166 & 21.64 & 0.1942 & 230.86 & \cellcolor{lightorange}0.8874 & \cellcolor{lightorange}25.71 & 0.1314 & \cellcolor{lightorange}388.02\\
GDAGS & 0.8129 & 21.53 & 0.2068 & \cellcolor{lightorange}223.96 & 0.8858 & 25.49 & 0.1380 & 415.07\\
Pixel-GS & \cellcolor{lightred}0.8279 & \cellcolor{lightred}22.13 & \cellcolor{lightred}0.1780 & 906.22 & 0.8868 & 25.48 & \cellcolor{lightred}0.1203 & 1228.62\\
ACD-GS & 0.8103 & 22.01 & 0.2298 & \cellcolor{lightred}123.27 & 0.8813 & 25.51 & 0.1434 & \cellcolor{lightred}177.46\\
\midrule
& \multicolumn{4}{c}{\textbf{Bicycle}} & \multicolumn{4}{c}{\textbf{Stump}} \\
\cmidrule(lr){2-5} \cmidrule(lr){6-9}
Method & SSIM$\uparrow$ & PSNR$\uparrow$ & LPIPS$\downarrow$ & Size(MB)$\downarrow$ & SSIM$\uparrow$ & PSNR$\uparrow$ & LPIPS$\downarrow$ & Size(MB)$\downarrow$ \\
\midrule
3DGS & 0.7666 & 25.28 & 0.2081 & \cellcolor{lightorange}1154.55 & 0.7714 & 26.66 & 0.2163 & 1026.20\\
Mip-Splatting & \cellcolor{lightred}0.7924 & \cellcolor{lightred}25.55 & \cellcolor{lightred}0.1669 & 1977.66 & \cellcolor{lightred}0.7902 & \cellcolor{lightorange}26.89 & \cellcolor{lightorange}0.1883 & 1440.26\\
Mini-Splatting & 0.7165 & 23.57 & 0.2508 & 1425.72 & 0.7601 & 26.03 & 0.2282 & 1274.58\\
AbsGS & 0.7817 & 25.26 & 0.1720 & 1411.36 & 0.7768 & 26.66 & 0.1967 & 1106.18\\
GDAGS & \cellcolor{lightorange}0.7899 & 25.42 & \cellcolor{lightorange}0.1696 & 1437.29 & 0.7743 & \cellcolor{lightred}26.96 & \cellcolor{lightred}0.1874 & \cellcolor{lightorange}949.72\\
Pixel-GS & 0.7617 & 25.26 & 0.1720 & 2142.21 & 0.7768 & 26.66 & 0.1967 & 1618.47\\
ACD-GS & 0.7861 & \cellcolor{lightorange}25.47 & 0.1799 & \cellcolor{lightred}543.99 & \cellcolor{lightorange}0.7856 & 26.83 & 0.2051 & \cellcolor{lightred}318.65\\
\midrule
& \multicolumn{4}{c}{\textbf{Treehill}} & \multicolumn{4}{c}{\textbf{Mean}} \\
\cmidrule(lr){2-5} \cmidrule(lr){6-9}
Method & SSIM$\uparrow$ & PSNR$\uparrow$ & LPIPS$\downarrow$ & Size(MB)$\downarrow$ & SSIM$\uparrow$ & PSNR$\uparrow$ & LPIPS$\downarrow$ & Size(MB)$\downarrow$ \\
\midrule
3DGS & 0.6335 & \cellcolor{lightred}22.49 & 0.3251 & 770.88 & 0.7754 & 24.39 & 0.2175 & \cellcolor{lightorange}739.56\\
Mip-Splatting & \cellcolor{lightorange}0.6387 & 22.34 & \cellcolor{lightred}0.2744 & 1223.14 & \cellcolor{lightred}0.7882 & \cellcolor{lightred}24.48 & \cellcolor{lightred}0.1884 & 1156.04\\
Mini-Splatting & 0.6244 & 22.06 & 0.3254 & \cellcolor{lightred}135.32 & 0.7336 & 22.82 & 0.2572 & 971.51\\
AbsGS & 0.6139 & 21.96 & 0.2795 & 1221.96 & 0.7753 & 24.25 & 0.1948 & 871.68\\
GDAGS & 0.6246 & 22.06 & \cellcolor{lightorange}0.2786 & 1171.14 & 0.7775 & 24.29 & 0.1961 & 839.44\\
Pixel-GS & 0.6139 & 21.96 & 0.2795 & 1868.62 & 0.7734 & 24.30 & \cellcolor{lightorange}0.1893 & 1552.83\\
ACD-GS & \cellcolor{lightred}0.6457 & \cellcolor{lightorange}22.36 & 0.2936 & \cellcolor{lightorange}472.37 & \cellcolor{lightorange}0.7818 & \cellcolor{lightorange}24.44 & 0.2104 & \cellcolor{lightred}327.15\\
\bottomrule
\end{tabular}
\end{table}

%% file: tables/efficiency_v8.tex
\begin{table}[t]
\centering
\caption{Efficiency Metrics Comparison}
\label{tab:efficiency_summary}
\label{tab:effiency}
\footnotesize
\setlength{\tabcolsep}{1pt}
\begin{tabular}{@{}lccccccccc@{}}
\toprule
& \multicolumn{3}{c}{\textbf{Deep Blending}} & \multicolumn{3}{c}{\textbf{Mip-NeRF 360}} & \multicolumn{3}{c}{\textbf{Tanks \& Temples}} \\
\cmidrule(lr){2-4} \cmidrule(lr){5-7} \cmidrule(lr){8-10}
Method & Time(s)$\downarrow$ & Size(MB)$\downarrow$ & Points(M)$\downarrow$ & Time(s)$\downarrow$ & Size(MB)$\downarrow$ & Points(M)$\downarrow$ & Time(s)$\downarrow$ & Size(MB)$\downarrow$ & Points(M)$\downarrow$ \\
\midrule
3DGS & 842 & 367.30 & \cellcolor{lightorange}1.5500 & 683 & \cellcolor{lightorange}638.40 & \cellcolor{lightorange}2.7000 & 552 & 373.10 & 1.5800 \\
Mip-Splatting & 1007 & 563.20 & 2.3400 & 956 & 991.10 & 4.1200 & 716 & 569.60 & 2.3700 \\
Mini-Splatting & 1125 & 959.10 & 4.1300 & 893 & 1108.60 & 4.6900 & 857 & 1011.00 & 4.2700 \\
AbsGS & 795 & 400.40 & 1.6900 & 683 & 716.40 & 3.0300 & 433 & \cellcolor{lightorange}309.40 & \cellcolor{lightorange}1.3100 \\
GDAGS & \cellcolor{lightorange}628 & \cellcolor{lightorange}360.60 & 1.5800 & \cellcolor{lightorange}517 & 707.10 & 2.9900 & \cellcolor{lightorange}350 & 319.50 & 1.3500 \\
Pixel-GS & 978 & 694.80 & 2.9400 & 1083 & 1315.10 & 5.5600 & 946 & 1067.40 & 4.5100 \\
ACD-GS & \cellcolor{lightred}545 & \cellcolor{lightred}151.90 & \cellcolor{lightred}0.6420 & \cellcolor{lightred}345 & \cellcolor{lightred}274.30 & \cellcolor{lightred}1.1600 & \cellcolor{lightred}333 & \cellcolor{lightred}150.40 & \cellcolor{lightred}0.6357 \\
\bottomrule
\end{tabular}
\end{table}

%% file: tables/alpha_beta_ablation_v8.tex
\begin{table}[t]
\centering
\begin{minipage}[t]{0.48\linewidth}
  \centering
  \caption{Comparison of different settings for $\alpha$ and $\beta$, varying one parameter at a time.}
  \label{tab:parametre10}
  \footnotesize
  \setlength{\tabcolsep}{2pt}
  \begin{tabular}{@{}lccccc@{}}
    \toprule
    $\alpha$ & $\beta$ & PSNR$\uparrow$ & SSIM$\uparrow$ & LPIPS$\downarrow$ & Size(MB)$\downarrow$ \\
    \midrule
    0.5 & 1.5 & 25.5890 & 0.8424 & 0.2387 & 161.37 \\
    0.8 & 1.5 & 25.5913 & 0.8450 & \textbf{0.2363} & \textbf{151.85} \\
    1.0 & 1.5 & \textbf{25.8259} & \textbf{0.8469} & 0.2520 & 162.54 \\
    \midrule
    0.8 & 1.0 & 25.4263 & 0.8412 & 0.2451 & 157.35 \\
    0.8 & 2.5 & 25.4441 & 0.8415 & 0.2436 & 164.87 \\
    \bottomrule
  \end{tabular}
\end{minipage}\hfill
\begin{minipage}[t]{0.48\linewidth}
  \centering
  \caption{Quantitative ablation on DeepBlending. CAD, MPG, and CAO denote the three components.}
  \label{tab:ablation}
  \footnotesize
  \setlength{\tabcolsep}{2pt}
  \begin{tabular}{@{}lcccccc@{}}
    \toprule
    CAD & MPG & CAO & Size(MB)$\downarrow$ & PSNR$\uparrow$ & SSIM$\uparrow$ & LPIPS$\downarrow$ \\
    \midrule
    $\times$ & $\times$ & $\times$ & 360.60 & 25.22 & 0.838 & 0.225 \\
    $\checkmark$ & $\times$ & $\times$ & 302.40 & 25.38 & 0.840 & 0.223 \\
    $\checkmark$ & $\checkmark$ & $\times$ & 159.70 & 25.38 & 0.841 & \textbf{0.221} \\
    $\checkmark$ & $\checkmark$ & $\checkmark$ & \textbf{151.90} & \textbf{25.59} & \textbf{0.845} & 0.236 \\
    \bottomrule
  \end{tabular}
\end{minipage}
\end{table}

%% file: sensitivity_v8.tex
\subsection{Sensitivity Analysis}
\label{sec:views_quota}

\begin{table}[t]
  \centering
  \caption{Sensitivity to the number of auxiliary views $K$ and quota ratio $\mu$.}
  \label{tab:views_quota}
  \scriptsize
  \begin{tabular}{@{}lccccc@{}}
    \toprule
    Setting & PSNR$\uparrow$ & SSIM$\uparrow$ & LPIPS$\downarrow$ & Points$\downarrow$ & Size(MB)$\downarrow$ \\
    \midrule
$K=4$ & 25.3380 & 0.8432 & 0.2393 & \textbf{550,173} & \textbf{136.44}\\
$K=8$ & \textbf{25.5914} & \textbf{0.8450} & \textbf{0.2363} & 642,045 & 151.85\\
$K=12$ & 25.3605 & 0.8440 & 0.2390 & 661,947 & 156.88\\
    \midrule
$\mu=0.004$ & 25.5145 & 0.8444 & \textbf{0.2363} & \textbf{641,737} & \textbf{151.78}\\
$\mu=0.008$ & \textbf{25.5913} & \textbf{0.8450} & \textbf{0.2363} & 642,044 & 151.85\\
$\mu=0.012$ & 25.4211 & 0.8432 & 0.2372 & 642,241 & 151.90\\
    \bottomrule
  \end{tabular}
\end{table}

We additionally evaluate the number of auxiliary views $K$ and the candidate quota ratio $\mu$ on the complete Deep Blending dataset. Increasing $K$ from 4 to 8 improves PSNR, SSIM, and LPIPS, but retains more Gaussians; increasing $K$ to 12 provides no consistent quality benefit and increases the model size, so we use $K=8$. For the quota ratio, $\mu=0.004$ yields a marginally smaller model, whereas $\mu=0.008$ achieves higher PSNR and SSIM with the same LPIPS; $\mu=0.012$ provides no additional quality improvement. We therefore use $\mu=0.008$ as the default.

%% file: 05Conclusion_v8.tex
\vspace{-0.8em}
\section{Discussion and Limitations}
\label{sec:discussion}
While ACD-GS achieves significant native sparsification, extreme compression inevitably degrades high-frequency texture fidelity due to the inherent capacity limits of pure explicit representations. To address this bottleneck, future work could explore hybrid architectures, replacing explicit spherical harmonics with lightweight Multi-Layer Perceptrons (MLPs) to maintain rendering quality under stricter point budgets. Furthermore, because ACD-GS operates orthogonally to post-hoc compression pipelines, seamlessly integrating it with downstream quantization and entropy coding frameworks presents a promising direction to unlock ultra-compact assets for low-bandwidth streaming.


\vspace{-0.8em}
\section{Conclusion}
\label{sec:conclusion}


We present ACD-GS, an Asymmetric Curvature-aware Densification framework for compact 3D Gaussian Splatting. Second-order depth priors decouple cloning and splitting, suppressing unnecessary cloning in planar regions while promoting splitting along geometric boundaries. A multi-view photometric gate filters transient noise, and adaptive thresholding with lifecycle-aware pruning regulates capacity. Experiments show that ACD-GS reduces Gaussian count and storage by nearly 60\% across benchmarks while preserving competitive rendering fidelity. By directly preventing redundancy during training, ACD-GS offers a practical, memory-efficient foundation for scalable 3D Gaussian representations.

%% file: appendix_v8.tex
\clearpage
\appendix
\section{Hyperparameter Sensitivity Analysis}
\label{sec:hyperparameter_sensitivity}

We further evaluate the sensitivity of ACD-GS to three key hyperparameters, including the soft threshold $\tau$, the capacity-control parameters $(\alpha,\beta)$, and the rate of lifecycle-aware opacity pruning. As shown in Tables~\ref{tab:soft_threshold_combined}--\ref{tab:lifecycle_combined}, ACD-GS exhibits stable rendering quality across different hyperparameter settings, while the representation size and the number of Gaussians vary more noticeably. For the soft threshold $\tau$, all tested settings achieve comparable average PSNR, SSIM, and LPIPS, indicating that the proposed soft modulation is not overly sensitive to the exact threshold value. The default setting $\tau=0.15$ provides a balanced trade-off, maintaining competitive rendering quality while yielding fewer Gaussians on average than the alternatives.

For the capacity-control parameters $(\alpha,\beta)$, the default configuration $(0.8,1.5)$ consistently produces the most compact representations across almost all scenes. Compared with both smaller and larger parameter settings, it substantially reduces the average storage and Gaussian count while preserving the best average PSNR and SSIM. This suggests that moderate capacity regulation is important: overly conservative control may retain redundant primitives, whereas overly aggressive densification-pruning schedules can increase representation size without clear quality gains.

Finally, varying the lifecycle-aware pruning rate leads to only minor changes in reconstruction quality. The average metrics remain very close across different rates, demonstrating that the pruning schedule is robust within a reasonable range. Although a smaller rate can slightly reduce the average model size, the default rate achieves the best overall PSNR, SSIM, and LPIPS, indicating a favorable balance between compactness and fidelity. Overall, these results show that ACD-GS is robust to hyperparameter variations and that the chosen default configuration provides a stable quality-efficiency trade-off across diverse scenes.

\setcounter{table}{0}
\renewcommand{\thetable}{A\arabic{table}}

\input{tables/appendix_combined_01_v8}
\input{tables/appendix_combined_02_v8}
\input{tables/appendix_combined_03_v8}

\section{Detailed Results}
\label{sec:detailed_results}
We provide detailed per-scene results on three benchmark datasets in the appendix. Specifically, Tables~\ref{tab:deepblending_combined}, \ref{tab:mipnerf360_combined}, and \ref{tab:tandt_combined} report per-scene rendering quality together with model size, Gaussian count, and training time for Deep Blending, Mip-NeRF 360, and Tanks \& Temples, respectively.

\input{tables/appendix_combined_04_v8}
\input{tables/appendix_combined_05_v8}
\input{tables/appendix_combined_06_v8}

\section{More Visual comparison}
\begin{figure}[H]
    \centering
    \includegraphics[width=\linewidth]{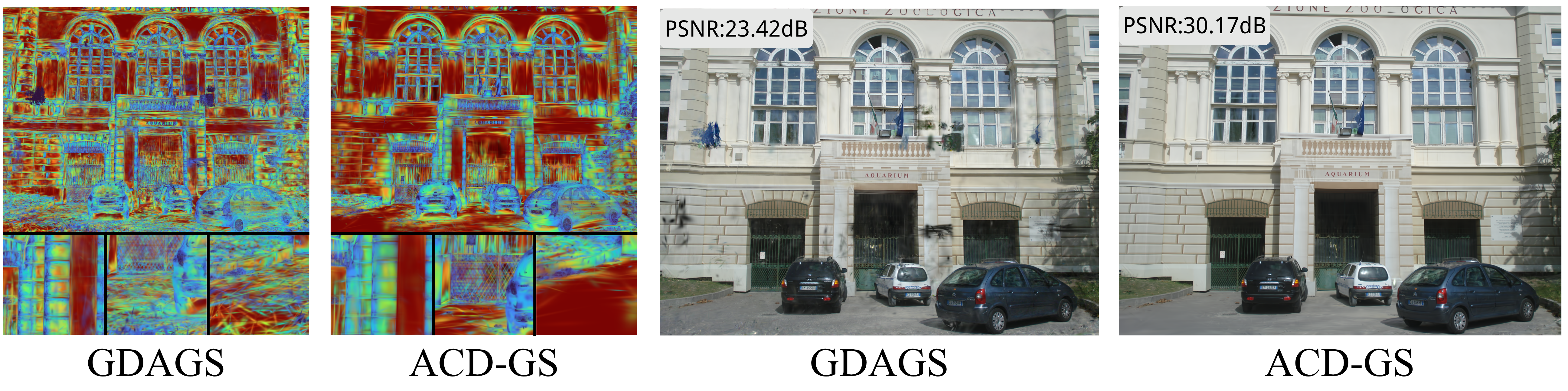}
    \caption{
    Qualitative comparison of Gaussian distributions.
    Our curvature-aware strategy allocates sparse and elongated Gaussians to smooth surfaces,
    while preserving sharp boundaries with compact fine-scale primitives.
    }
    \label{fig:observation}
\end{figure}

\input{geometry_v8}

%% file: tables/appendix_combined_01_v8.tex
\begingroup\small
\begin{longtable}{@{}llccccc@{}}
\caption{Quality and efficiency under different depth soft thresholds.}\label{tab:soft_threshold_combined}\\
\toprule
Scene & Setting & SSIM$\uparrow$ & PSNR$\uparrow$ & LPIPS$\downarrow$ & Size(MB)$\downarrow$ & Points(M)$\downarrow$ \\
\midrule
\endfirsthead
\multicolumn{7}{l}{\tablename~\thetable{} (continued)}\\
\toprule
Scene & Setting & SSIM$\uparrow$ & PSNR$\uparrow$ & LPIPS$\downarrow$ & Size(MB)$\downarrow$ & Points(M)$\downarrow$ \\
\midrule
\endhead
\midrule
\multicolumn{7}{r}{Continued on next page}\\
\endfoot
\bottomrule
\endlastfoot
Aquarium-20 & $\tau=0.05$ & 0.8722 & 26.16 & 0.1673 & 52.04 & 0.2777 \\*
 & $\tau=0.15$ & \textbf{0.8738} & \textbf{26.42} & \textbf{0.1655} & \textbf{41.76} & \textbf{0.1766} \\*
 & $\tau=0.25$ & 0.8726 & 26.03 & 0.1656 & 51.73 & 0.2764 \\
\midrule
Bedroom & $\tau=0.05$ & 0.9070 & 29.00 & \textbf{0.2897} & 121.89 & 0.5731 \\*
 & $\tau=0.15$ & 0.9073 & 29.04 & 0.2899 & \textbf{111.74} & \textbf{0.4724} \\*
 & $\tau=0.25$ & \textbf{0.9076} & \textbf{29.04} & 0.2903 & 121.59 & 0.5718 \\
\midrule
Boats & $\tau=0.05$ & 0.9101 & 26.41 & 0.2209 & 167.23 & 0.7648 \\*
 & $\tau=0.15$ & 0.9087 & 26.15 & 0.2213 & \textbf{158.20} & \textbf{0.6689} \\*
 & $\tau=0.25$ & \textbf{0.9102} & \textbf{26.47} & \textbf{0.2197} & 163.70 & 0.7499 \\
\midrule
Bridge & $\tau=0.05$ & 0.7873 & 24.09 & \textbf{0.3102} & 163.37 & 0.7484 \\*
 & $\tau=0.15$ & \textbf{0.7904} & \textbf{24.18} & 0.3104 & \textbf{149.19} & \textbf{0.6308} \\*
 & $\tau=0.25$ & 0.7897 & 23.91 & 0.3103 & 161.57 & 0.7409 \\
\midrule
CreepyAttic & $\tau=0.05$ & 0.8903 & 30.62 & 0.2457 & 177.24 & 0.8071 \\*
 & $\tau=0.15$ & \textbf{0.8908} & \textbf{30.70} & \textbf{0.2442} & \textbf{165.98} & \textbf{0.7018} \\*
 & $\tau=0.25$ & 0.8904 & 30.65 & 0.2453 & 175.90 & 0.8014 \\
\midrule
Hugo-1 & $\tau=0.05$ & 0.8464 & \textbf{24.53} & 0.1613 & 124.61 & 0.5846 \\*
 & $\tau=0.15$ & \textbf{0.8468} & 24.45 & \textbf{0.1608} & \textbf{110.88} & \textbf{0.4688} \\*
 & $\tau=0.25$ & 0.8423 & 23.92 & 0.1637 & 122.62 & 0.5762 \\
\midrule
Library & $\tau=0.05$ & 0.7814 & 23.98 & 0.3597 & 254.96 & 1.1357 \\*
 & $\tau=0.15$ & 0.7798 & \textbf{24.10} & 0.3592 & \textbf{245.24} & \textbf{1.0369} \\*
 & $\tau=0.25$ & \textbf{0.7820} & 24.10 & \textbf{0.3592} & 256.48 & 1.1421 \\
\midrule
Lumber & $\tau=0.05$ & 0.7842 & 22.28 & 0.2653 & 508.99 & 2.2098 \\*
 & $\tau=0.15$ & \textbf{0.7867} & \textbf{22.46} & \textbf{0.2634} & 495.92 & 2.0968 \\*
 & $\tau=0.25$ & 0.7843 & 22.34 & 0.2661 & \textbf{466.31} & \textbf{2.0293} \\
\midrule
Museum-1 & $\tau=0.05$ & 0.8445 & \textbf{23.64} & 0.1469 & \textbf{198.21} & \textbf{0.8958} \\*
 & $\tau=0.15$ & \textbf{0.8450} & 23.49 & \textbf{0.1452} & 223.28 & 0.9440 \\*
 & $\tau=0.25$ & 0.8442 & 23.45 & 0.1456 & 234.51 & 1.0493 \\
\midrule
Museum-2 & $\tau=0.05$ & \textbf{0.9013} & \textbf{28.07} & 0.1065 & \textbf{108.91} & 0.5182 \\*
 & $\tau=0.15$ & 0.9002 & 27.61 & \textbf{0.1056} & 121.68 & \textbf{0.5145} \\*
 & $\tau=0.25$ & 0.8987 & 27.55 & 0.1066 & 132.65 & 0.6186 \\
\midrule
NightSnow & $\tau=0.05$ & 0.9055 & 30.16 & 0.2242 & \textbf{61.68} & 0.3185 \\*
 & $\tau=0.15$ & 0.9054 & 30.22 & 0.2200 & 64.62 & \textbf{0.2732} \\*
 & $\tau=0.25$ & \textbf{0.9062} & \textbf{30.33} & \textbf{0.2199} & 73.96 & 0.3704 \\
\midrule
Playroom & $\tau=0.05$ & 0.8949 & 29.67 & 0.2921 & \textbf{112.51} & \textbf{0.5334} \\*
 & $\tau=0.15$ & \textbf{0.8960} & \textbf{30.07} & \textbf{0.2883} & 129.12 & 0.5459 \\*
 & $\tau=0.25$ & 0.8952 & 29.92 & 0.2888 & 137.39 & 0.6386 \\
\midrule
Ponche & $\tau=0.05$ & 0.8211 & 23.11 & 0.2558 & \textbf{103.84} & 0.4968 \\*
 & $\tau=0.15$ & 0.8191 & 23.07 & 0.2537 & 114.83 & \textbf{0.4855} \\*
 & $\tau=0.25$ & \textbf{0.8234} & \textbf{23.24} & \textbf{0.2494} & 123.42 & 0.5796 \\
\midrule
SaintAnne & $\tau=0.05$ & 0.8195 & \textbf{25.60} & 0.3372 & \textbf{135.70} & \textbf{0.6315} \\*
 & $\tau=0.15$ & \textbf{0.8203} & 25.51 & 0.3309 & 160.94 & 0.6805 \\*
 & $\tau=0.25$ & 0.8189 & 25.34 & \textbf{0.3309} & 171.86 & 0.7844 \\
\midrule
Shed & $\tau=0.05$ & 0.9338 & 28.32 & 0.1956 & \textbf{96.72} & 0.4666 \\*
 & $\tau=0.15$ & 0.9295 & 27.28 & 0.1963 & 107.75 & \textbf{0.4556} \\*
 & $\tau=0.25$ & \textbf{0.9347} & \textbf{28.39} & \textbf{0.1912} & 117.14 & 0.5530 \\
\midrule
Tree-18 & $\tau=0.05$ & 0.8027 & 22.18 & 0.1747 & \textbf{81.61} & 0.4028 \\*
 & $\tau=0.15$ & \textbf{0.8052} & 22.22 & \textbf{0.1650} & 84.78 & \textbf{0.3584} \\*
 & $\tau=0.25$ & 0.8044 & \textbf{22.33} & 0.1658 & 95.18 & 0.4602 \\
\midrule
Yellowhouse-12 & $\tau=0.05$ & 0.6536 & 17.58 & 0.3066 & \textbf{90.42} & 0.4400 \\*
 & $\tau=0.15$ & \textbf{0.6608} & \textbf{18.09} & 0.2975 & 95.58 & \textbf{0.4041} \\*
 & $\tau=0.25$ & 0.6485 & 17.06 & \textbf{0.2972} & 109.54 & 0.5209 \\
\midrule
Mean & $\tau=0.05$ & 0.8445 & \textbf{25.61} & 0.2388 & \textbf{150.58} & 0.6944 \\*
 & $\tau=0.15$ & \textbf{0.8451} & 25.59 & 0.2363 & 151.85 & \textbf{0.6420} \\*
 & $\tau=0.25$ & 0.8113 & 25.53 & \textbf{0.2362} & 151.43 & 0.7331 \\
\midrule
\end{longtable}
\endgroup

%% file: tables/appendix_combined_02_v8.tex
\begingroup\small
\begin{longtable}{@{}llccccc@{}}
\caption{Quality and efficiency under different settings of $\alpha$ and $\beta$.}\label{tab:alpha_beta_combined}\\
\toprule
Scene & Setting & SSIM$\uparrow$ & PSNR$\uparrow$ & LPIPS$\downarrow$ & Size(MB)$\downarrow$ & Points(M)$\downarrow$ \\
\midrule
\endfirsthead
\multicolumn{7}{l}{\tablename~\thetable{} (continued)}\\
\toprule
Scene & Setting & SSIM$\uparrow$ & PSNR$\uparrow$ & LPIPS$\downarrow$ & Size(MB)$\downarrow$ & Points(M)$\downarrow$ \\
\midrule
\endhead
\midrule
\multicolumn{7}{r}{Continued on next page}\\
\endfoot
\bottomrule
\endlastfoot
Aquarium-20 & $(0.5,1.0)$ & 0.8517 & 23.38 & 0.1751 & 132.12 & 0.5586 \\*
 & $(0.8,1.5)$ & \textbf{0.8738} & \textbf{26.42} & \textbf{0.1655} & \textbf{41.76} & \textbf{0.1766} \\*
 & $(1.0,2.5)$ & 0.8486 & 24.40 & 0.1760 & 140.75 & 0.5951 \\
\midrule
Bedroom & $(0.5,1.0)$ & 0.9068 & 28.87 & 0.2729 & 185.58 & 0.7847 \\*
 & $(0.8,1.5)$ & \textbf{0.9073} & \textbf{29.04} & 0.2899 & \textbf{111.74} & \textbf{0.4724} \\*
 & $(1.0,2.5)$ & 0.9072 & 28.92 & \textbf{0.2719} & 197.59 & 0.8355 \\
\midrule
Boats & $(0.5,1.0)$ & 0.9122 & 26.57 & 0.2074 & 224.80 & 0.9505 \\*
 & $(0.8,1.5)$ & 0.9087 & 26.15 & 0.2213 & \textbf{158.20} & \textbf{0.6689} \\*
 & $(1.0,2.5)$ & \textbf{0.9136} & \textbf{26.72} & \textbf{0.2020} & 238.70 & 1.0092 \\
\midrule
Bridge & $(0.5,1.0)$ & 0.7830 & 23.86 & 0.3175 & 152.05 & 0.6429 \\*
 & $(0.8,1.5)$ & \textbf{0.7904} & \textbf{24.18} & \textbf{0.3104} & \textbf{149.19} & \textbf{0.6308} \\*
 & $(1.0,2.5)$ & 0.7823 & 23.41 & 0.3147 & 168.09 & 0.7107 \\
\midrule
CreepyAttic & $(0.5,1.0)$ & 0.8910 & 30.64 & 0.2176 & 463.91 & 1.9615 \\*
 & $(0.8,1.5)$ & 0.8908 & \textbf{30.70} & 0.2442 & \textbf{165.98} & \textbf{0.7018} \\*
 & $(1.0,2.5)$ & \textbf{0.8910} & 30.55 & \textbf{0.2166} & 483.30 & 2.0435 \\
\midrule
Hugo-1 & $(0.5,1.0)$ & 0.8319 & 23.87 & \textbf{0.1498} & 256.77 & 1.0857 \\*
 & $(0.8,1.5)$ & \textbf{0.8468} & \textbf{24.45} & 0.1608 & \textbf{110.88} & \textbf{0.4688} \\*
 & $(1.0,2.5)$ & 0.8276 & 23.61 & 0.1504 & 312.31 & 1.3205 \\
\midrule
Library & $(0.5,1.0)$ & 0.7722 & 23.84 & 0.3580 & 371.66 & 1.5714 \\*
 & $(0.8,1.5)$ & \textbf{0.7798} & \textbf{24.10} & 0.3592 & \textbf{245.24} & \textbf{1.0369} \\*
 & $(1.0,2.5)$ & 0.7705 & 23.86 & \textbf{0.3566} & 405.50 & 1.7145 \\
\midrule
Lumber & $(0.5,1.0)$ & 0.7899 & \textbf{22.63} & 0.2479 & 560.72 & 2.3708 \\*
 & $(0.8,1.5)$ & 0.7867 & 22.46 & 0.2634 & \textbf{495.92} & \textbf{2.0968} \\*
 & $(1.0,2.5)$ & \textbf{0.7924} & 22.56 & \textbf{0.2364} & 681.21 & 2.8802 \\
\midrule
Museum-1 & $(0.5,1.0)$ & 0.8348 & 22.81 & 0.1373 & 395.33 & 1.6715 \\*
 & $(0.8,1.5)$ & \textbf{0.8450} & \textbf{23.49} & 0.1452 & \textbf{223.28} & \textbf{0.9440} \\*
 & $(1.0,2.5)$ & 0.8354 & 22.82 & \textbf{0.1359} & 414.99 & 1.7546 \\
\midrule
Museum-2 & $(0.5,1.0)$ & 0.9016 & 26.62 & 0.0925 & 212.18 & 0.8971 \\*
 & $(0.8,1.5)$ & 0.9002 & \textbf{27.61} & 0.1056 & \textbf{121.68} & \textbf{0.5145} \\*
 & $(1.0,2.5)$ & \textbf{0.9022} & 27.15 & \textbf{0.0909} & 228.20 & 0.9648 \\
\midrule
NightSnow & $(0.5,1.0)$ & 0.9147 & 30.49 & 0.1912 & 94.44 & 0.3993 \\*
 & $(0.8,1.5)$ & 0.9054 & 30.22 & 0.2200 & \textbf{64.62} & \textbf{0.2732} \\*
 & $(1.0,2.5)$ & \textbf{0.9151} & \textbf{30.51} & \textbf{0.1888} & 101.40 & 0.4287 \\
\midrule
Playroom & $(0.5,1.0)$ & \textbf{0.8960} & \textbf{30.18} & 0.2675 & 287.29 & 1.2147 \\*
 & $(0.8,1.5)$ & 0.8960 & 30.07 & 0.2883 & \textbf{129.12} & \textbf{0.5459} \\*
 & $(1.0,2.5)$ & 0.8954 & 30.07 & \textbf{0.2671} & 300.31 & 1.2697 \\
\midrule
Ponche & $(0.5,1.0)$ & \textbf{0.8194} & 22.54 & \textbf{0.2289} & 181.75 & 0.7684 \\*
 & $(0.8,1.5)$ & 0.8191 & \textbf{23.07} & 0.2537 & \textbf{114.83} & \textbf{0.4855} \\*
 & $(1.0,2.5)$ & 0.8148 & 21.89 & 0.2291 & 196.96 & 0.8328 \\
\midrule
SaintAnne & $(0.5,1.0)$ & 0.8225 & 25.61 & 0.3109 & 199.34 & 0.8428 \\*
 & $(0.8,1.5)$ & 0.8203 & 25.51 & 0.3309 & \textbf{160.94} & \textbf{0.6805} \\*
 & $(1.0,2.5)$ & \textbf{0.8236} & \textbf{25.68} & \textbf{0.2955} & 238.51 & 1.0084 \\
\midrule
Shed & $(0.5,1.0)$ & 0.9484 & 32.21 & 0.1536 & 168.72 & 0.7134 \\*
 & $(0.8,1.5)$ & 0.9295 & 27.28 & 0.1963 & \textbf{107.75} & \textbf{0.4556} \\*
 & $(1.0,2.5)$ & \textbf{0.9489} & \textbf{32.37} & \textbf{0.1510} & 174.77 & 0.7389 \\
\midrule
Tree-18 & $(0.5,1.0)$ & 0.7722 & 20.85 & 0.2055 & 164.06 & 0.6937 \\*
 & $(0.8,1.5)$ & \textbf{0.8052} & 22.22 & \textbf{0.1650} & \textbf{84.78} & \textbf{0.3584} \\*
 & $(1.0,2.5)$ & 0.7873 & \textbf{22.28} & 0.1778 & 200.11 & 0.8461 \\
\midrule
Yellowhouse-12 & $(0.5,1.0)$ & 0.6240 & 15.33 & 0.3198 & 154.89 & 0.6549 \\*
 & $(0.8,1.5)$ & \textbf{0.6608} & \textbf{18.09} & \textbf{0.2975} & \textbf{95.58} & \textbf{0.4041} \\*
 & $(1.0,2.5)$ & 0.6397 & 17.16 & 0.3077 & 145.98 & 0.6172 \\
\midrule
Mean & $(0.5,1.0)$ & 0.8395 & 25.31 & 0.2267 & 247.39 & 1.0460 \\*
 & $(0.8,1.5)$ & \textbf{0.8451} & \textbf{25.59} & 0.2363 & \textbf{151.85} & \textbf{0.6420} \\*
 & $(1.0,2.5)$ & 0.8409 & 25.53 & \textbf{0.2217} & 272.28 & 1.1512 \\
\midrule
\end{longtable}
\endgroup

%% file: tables/appendix_combined_03_v8.tex
\begingroup\small
\begin{longtable}{@{}llccccc@{}}
\caption{Quality and efficiency under different opacity-pruning rates.}\label{tab:lifecycle_combined}\\
\toprule
Scene & Setting & SSIM$\uparrow$ & PSNR$\uparrow$ & LPIPS$\downarrow$ & Size(MB)$\downarrow$ & Points(M)$\downarrow$ \\
\midrule
\endfirsthead
\multicolumn{7}{l}{\tablename~\thetable{} (continued)}\\
\toprule
Scene & Setting & SSIM$\uparrow$ & PSNR$\uparrow$ & LPIPS$\downarrow$ & Size(MB)$\downarrow$ & Points(M)$\downarrow$ \\
\midrule
\endhead
\midrule
\multicolumn{7}{r}{Continued on next page}\\
\endfoot
\bottomrule
\endlastfoot
Aquarium-20 & Rate $=0.5$ & \textbf{0.8739} & 26.20 & 0.1652 & \textbf{40.21} & \textbf{0.1700} \\*
 & Ours (Rate $=1.0$) & 0.8738 & \textbf{26.42} & 0.1655 & 41.76 & 0.1766 \\*
 & Rate $=1.5$ & 0.8729 & 26.28 & \textbf{0.1647} & 42.01 & 0.1776 \\
\midrule
Bedroom & Rate $=0.5$ & 0.9078 & \textbf{29.10} & 0.2898 & \textbf{111.66} & \textbf{0.4721} \\*
 & Ours (Rate $=1.0$) & 0.9073 & 29.04 & 0.2899 & 111.74 & 0.4724 \\*
 & Rate $=1.5$ & \textbf{0.9084} & 29.07 & \textbf{0.2891} & 111.76 & 0.4725 \\
\midrule
Boats & Rate $=0.5$ & \textbf{0.9105} & \textbf{26.37} & \textbf{0.2207} & 158.79 & 0.6714 \\*
 & Ours (Rate $=1.0$) & 0.9087 & 26.15 & 0.2213 & 158.20 & 0.6689 \\*
 & Rate $=1.5$ & 0.9093 & 26.25 & 0.2222 & \textbf{153.99} & \textbf{0.6511} \\
\midrule
Bridge & Rate $=0.5$ & 0.7887 & 24.16 & 0.3110 & 153.40 & 0.6486 \\*
 & Ours (Rate $=1.0$) & \textbf{0.7904} & \textbf{24.18} & \textbf{0.3104} & \textbf{149.19} & \textbf{0.6308} \\*
 & Rate $=1.5$ & 0.7888 & 23.89 & 0.3107 & 150.39 & 0.6358 \\
\midrule
CreepyAttic & Rate $=0.5$ & 0.8902 & 30.60 & 0.2449 & 167.08 & 0.7064 \\*
 & Ours (Rate $=1.0$) & \textbf{0.8908} & 30.70 & \textbf{0.2442} & \textbf{165.98} & \textbf{0.7018} \\*
 & Rate $=1.5$ & 0.8902 & \textbf{30.71} & 0.2455 & 167.41 & 0.7078 \\
\midrule
Hugo-1 & Rate $=0.5$ & 0.8454 & 24.51 & 0.1621 & 111.38 & 0.4709 \\*
 & Ours (Rate $=1.0$) & \textbf{0.8468} & 24.45 & \textbf{0.1608} & \textbf{110.88} & \textbf{0.4688} \\*
 & Rate $=1.5$ & 0.8451 & \textbf{24.65} & 0.1620 & 114.41 & 0.4837 \\
\midrule
Library & Rate $=0.5$ & 0.7775 & 23.98 & 0.3600 & \textbf{244.56} & \textbf{1.0340} \\*
 & Ours (Rate $=1.0$) & 0.7798 & \textbf{24.10} & \textbf{0.3592} & 245.24 & 1.0369 \\*
 & Rate $=1.5$ & \textbf{0.7818} & 23.96 & 0.3601 & 247.66 & 1.0471 \\
\midrule
Lumber & Rate $=0.5$ & 0.7775 & 21.81 & 0.2704 & \textbf{469.24} & \textbf{1.9840} \\*
 & Ours (Rate $=1.0$) & \textbf{0.7867} & \textbf{22.46} & \textbf{0.2634} & 495.92 & 2.0968 \\*
 & Rate $=1.5$ & 0.7808 & 21.99 & 0.2670 & 480.98 & 2.0336 \\
\midrule
Museum-1 & Rate $=0.5$ & 0.8435 & 23.41 & 0.1469 & 223.95 & 0.9469 \\*
 & Ours (Rate $=1.0$) & \textbf{0.8450} & 23.49 & \textbf{0.1452} & \textbf{223.28} & \textbf{0.9440} \\*
 & Rate $=1.5$ & 0.8443 & \textbf{23.55} & 0.1457 & 223.91 & 0.9467 \\
\midrule
Museum-2 & Rate $=0.5$ & 0.8996 & 27.56 & 0.1061 & \textbf{120.31} & \textbf{0.5087} \\*
 & Ours (Rate $=1.0$) & \textbf{0.9002} & \textbf{27.61} & \textbf{0.1056} & 121.68 & 0.5145 \\*
 & Rate $=1.5$ & 0.8978 & 27.58 & 0.1082 & 120.43 & 0.5092 \\
\midrule
NightSnow & Rate $=0.5$ & 0.9034 & 29.98 & 0.2216 & \textbf{63.16} & \textbf{0.2671} \\*
 & Ours (Rate $=1.0$) & \textbf{0.9054} & 30.22 & 0.2200 & 64.62 & 0.2732 \\*
 & Rate $=1.5$ & 0.9053 & \textbf{30.45} & \textbf{0.2199} & 65.45 & 0.2767 \\
\midrule
Playroom & Rate $=0.5$ & \textbf{0.8962} & \textbf{30.12} & 0.2888 & \textbf{127.31} & \textbf{0.5383} \\*
 & Ours (Rate $=1.0$) & 0.8960 & 30.07 & \textbf{0.2883} & 129.12 & 0.5459 \\*
 & Rate $=1.5$ & 0.8957 & 29.93 & 0.2891 & 129.34 & 0.5469 \\
\midrule
Ponche & Rate $=0.5$ & \textbf{0.8236} & \textbf{23.22} & \textbf{0.2491} & \textbf{114.69} & \textbf{0.4849} \\*
 & Ours (Rate $=1.0$) & 0.8191 & 23.07 & 0.2537 & 114.83 & 0.4855 \\*
 & Rate $=1.5$ & 0.8223 & 23.09 & 0.2505 & 116.27 & 0.4916 \\
\midrule
SaintAnne & Rate $=0.5$ & 0.8209 & \textbf{25.74} & \textbf{0.3277} & 168.31 & 0.7116 \\*
 & Ours (Rate $=1.0$) & 0.8203 & 25.51 & 0.3309 & \textbf{160.94} & \textbf{0.6805} \\*
 & Rate $=1.5$ & \textbf{0.8216} & 25.66 & 0.3284 & 161.27 & 0.6819 \\
\midrule
Shed & Rate $=0.5$ & \textbf{0.9320} & \textbf{27.67} & \textbf{0.1943} & 106.29 & 0.4494 \\*
 & Ours (Rate $=1.0$) & 0.9295 & 27.28 & 0.1963 & 107.75 & 0.4556 \\*
 & Rate $=1.5$ & 0.9273 & 27.41 & 0.1999 & \textbf{104.45} & \textbf{0.4416} \\
\midrule
Tree-18 & Rate $=0.5$ & \textbf{0.8056} & \textbf{22.28} & 0.1656 & 87.19 & 0.3686 \\*
 & Ours (Rate $=1.0$) & 0.8052 & 22.22 & \textbf{0.1650} & \textbf{84.78} & \textbf{0.3584} \\*
 & Rate $=1.5$ & 0.8028 & 22.24 & 0.1678 & 92.55 & 0.3913 \\
\midrule
Yellowhouse-12 & Rate $=0.5$ & 0.6540 & 17.82 & 0.3058 & 96.52 & 0.4081 \\*
 & Ours (Rate $=1.0$) & \textbf{0.6608} & \textbf{18.09} & \textbf{0.2975} & \textbf{95.58} & \textbf{0.4041} \\*
 & Rate $=1.5$ & 0.6537 & 17.65 & 0.3009 & 97.89 & 0.4139 \\
\midrule
Mean & Rate $=0.5$ & 0.8441 & 25.56 & 0.2371 & \textbf{150.83} & \textbf{0.6377} \\*
 & Ours (Rate $=1.0$) & \textbf{0.8451} & \textbf{25.59} & \textbf{0.2363} & 151.85 & 0.6420 \\*
 & Rate $=1.5$ & 0.8440 & 25.55 & 0.2372 & 151.77 & 0.6417 \\
\midrule
\end{longtable}
\endgroup

%% file: tables/appendix_combined_04_v8.tex
\begingroup\small
\begin{longtable}{@{}llcccccc@{}}
\caption{Per-scene quality and efficiency on Deep Blending.}\label{tab:deepblending_combined}\\
\toprule
Scene & Method & SSIM$\uparrow$ & PSNR$\uparrow$ & LPIPS$\downarrow$ & Size(MB)$\downarrow$ & Points(M)$\downarrow$ & Time(s)$\downarrow$ \\
\midrule
\endfirsthead
\multicolumn{8}{l}{\tablename~\thetable{} (continued)}\\
\toprule
Scene & Method & SSIM$\uparrow$ & PSNR$\uparrow$ & LPIPS$\downarrow$ & Size(MB)$\downarrow$ & Points(M)$\downarrow$ & Time(s)$\downarrow$ \\
\midrule
\endhead
\midrule
\multicolumn{8}{r}{Continued on next page}\\
\endfoot
\bottomrule
\endlastfoot
Aquarium-20 & AbsGS & 0.8060 & 21.82 & 0.2288 & 183.16 & 0.7700 & 828 \\*
 & 3DGS & 0.8692 & 22.66 & 0.1607 & 177.44 & 0.7500 & 942 \\*
 & Mip-Splatting & 0.8507 & 22.52 & 0.1764 & 272.31 & 1.1300 & 1034 \\*
 & Pixel-GS & 0.8653 & 25.77 & \textbf{0.1582} & 268.57 & 1.1400 & 900 \\*
 & GDAGS & 0.8418 & 22.18 & 0.1865 & 159.86 & 0.7900 & 743 \\*
 & Mini-Splatting & 0.8509 & 23.81 & 0.1659 & 810.17 & 3.4300 & 1329 \\*
 & ACD-GS & \textbf{0.8738} & \textbf{26.42} & 0.1655 & \textbf{41.76} & \textbf{0.1800} & \textbf{649} \\
\midrule
Bedroom & AbsGS & 0.9012 & 28.85 & 0.2751 & 253.97 & 1.0700 & 854 \\*
 & 3DGS & 0.9072 & 28.86 & 0.2743 & 316.93 & 1.3400 & 691 \\*
 & Mip-Splatting & 0.9045 & 28.68 & 0.2744 & 473.69 & 1.9700 & 791 \\*
 & Pixel-GS & 0.9041 & 28.82 & 0.2748 & 719.19 & 3.0400 & 834 \\*
 & GDAGS & 0.9070 & \textbf{29.05} & \textbf{0.2701} & 283.22 & 1.2200 & 476 \\*
 & Mini-Splatting & 0.9007 & 28.28 & 0.2955 & 981.14 & 4.1500 & 976 \\*
 & ACD-GS & \textbf{0.9073} & 29.04 & 0.2899 & \textbf{111.74} & \textbf{0.4700} & \textbf{405} \\
\midrule
Boats & AbsGS & 0.9036 & 26.61 & 0.2162 & 399.18 & 1.6900 & 737 \\*
 & 3DGS & 0.9122 & 26.68 & 0.2099 & 389.69 & 1.6500 & 766 \\*
 & Mip-Splatting & \textbf{0.9159} & 26.75 & \textbf{0.1934} & 629.55 & 2.6200 & 930 \\*
 & Pixel-GS & 0.9145 & \textbf{26.98} & 0.1967 & 846.92 & 3.5800 & 975 \\*
 & GDAGS & 0.9137 & 26.67 & 0.2018 & 344.09 & 1.4800 & 568 \\*
 & Mini-Splatting & 0.9054 & 25.73 & 0.2133 & 1063.13 & 4.5000 & 1111 \\*
 & ACD-GS & 0.9087 & 26.15 & 0.2213 & \textbf{158.20} & \textbf{0.6700} & \textbf{509} \\
\midrule
Bridge & AbsGS & 0.7694 & 22.65 & 0.3123 & 245.33 & 1.0400 & 701 \\*
 & 3DGS & \textbf{0.8103} & 25.13 & 0.2884 & 259.22 & 1.1000 & 817 \\*
 & Mip-Splatting & 0.8062 & 24.41 & \textbf{0.2848} & 332.83 & 1.3800 & 921 \\*
 & Pixel-GS & 0.8056 & \textbf{25.77} & 0.3049 & 305.77 & 1.2900 & 780 \\*
 & GDAGS & 0.7834 & 23.59 & 0.3078 & 232.13 & 0.9900 & 593 \\*
 & Mini-Splatting & 0.7301 & 21.32 & 0.4274 & 952.32 & 4.0300 & 1269 \\*
 & ACD-GS & 0.7904 & 24.18 & 0.3104 & \textbf{149.19} & \textbf{0.6300} & \textbf{586} \\
\midrule
CreepyAttic & AbsGS & 0.8821 & 30.16 & 0.2199 & 704.01 & 2.9800 & 829 \\*
 & 3DGS & \textbf{0.8924} & 30.52 & 0.2245 & 758.95 & 3.2100 & 968 \\*
 & Mip-Splatting & 0.8860 & 30.16 & 0.2208 & 1142.25 & 4.7500 & 1235 \\*
 & Pixel-GS & 0.8841 & 30.24 & 0.2216 & 1558.26 & 6.5900 & 1329 \\*
 & GDAGS & 0.8890 & 30.43 & \textbf{0.2138} & 711.70 & 2.9400 & 596 \\*
 & Mini-Splatting & 0.8682 & 29.41 & 0.2800 & 1172.18 & 4.9600 & 1005 \\*
 & ACD-GS & 0.8908 & \textbf{30.70} & 0.2442 & \textbf{165.98} & \textbf{0.7000} & \textbf{422} \\
\midrule
Hugo-1 & AbsGS & 0.7891 & 22.58 & 0.1804 & 344.76 & 1.4600 & 811 \\*
 & 3DGS & 0.8429 & 23.85 & 0.1394 & 321.95 & 1.3600 & 830 \\*
 & Mip-Splatting & 0.8304 & 22.80 & 0.1486 & 494.73 & 2.0600 & 1030 \\*
 & Pixel-GS & 0.8355 & 23.43 & \textbf{0.1393} & 496.32 & 2.1000 & 923 \\*
 & GDAGS & 0.8208 & 23.39 & 0.1597 & 318.72 & 1.4800 & 699 \\*
 & Mini-Splatting & 0.8412 & 23.24 & 0.1433 & 873.89 & 3.6900 & 1075 \\*
 & ACD-GS & \textbf{0.8468} & \textbf{24.45} & 0.1608 & \textbf{110.88} & \textbf{0.4700} & \textbf{574} \\
\midrule
Library & AbsGS & 0.7311 & 23.49 & 0.4162 & 738.66 & 3.1200 & 991 \\*
 & 3DGS & \textbf{0.7806} & 23.87 & \textbf{0.3520} & 501.54 & 2.1200 & 900 \\*
 & Mip-Splatting & 0.7629 & 23.32 & 0.3576 & 829.34 & 3.4500 & 1144 \\*
 & Pixel-GS & 0.7767 & 23.73 & 0.3577 & 998.51 & 4.2200 & 1099 \\*
 & GDAGS & 0.7682 & 23.72 & 0.3570 & 568.79 & 2.5100 & 682 \\*
 & Mini-Splatting & 0.7616 & 23.33 & 0.3941 & 1129.81 & 4.7800 & 1216 \\*
 & ACD-GS & 0.7798 & \textbf{24.10} & 0.3592 & \textbf{245.24} & \textbf{1.0400} & \textbf{570} \\
\midrule
Lumber & AbsGS & 0.7691 & 22.04 & 0.2559 & 1163.96 & 4.9200 & 1188 \\*
 & 3DGS & 0.7705 & 22.03 & 0.2787 & 861.48 & 3.6400 & 1042 \\*
 & Mip-Splatting & \textbf{0.7947} & 22.61 & \textbf{0.2381} & 1411.98 & 5.8800 & 1442 \\*
 & Pixel-GS & 0.7777 & 22.24 & 0.2638 & 1549.63 & 6.5500 & 1414 \\*
 & GDAGS & 0.7907 & \textbf{22.67} & 0.2445 & 921.82 & 3.9700 & 755 \\*
 & Mini-Splatting & 0.7576 & 21.77 & 0.2694 & 1275.84 & 5.3900 & 1208 \\*
 & ACD-GS & 0.7867 & 22.46 & 0.2634 & \textbf{495.92} & \textbf{2.1000} & 623 \\
\midrule
Museum-1 & AbsGS & 0.7931 & 21.73 & 0.1689 & 568.56 & 2.4000 & 1005 \\*
 & 3DGS & 0.8364 & 23.32 & 0.1346 & 567.26 & 2.4000 & 1272 \\*
 & Mip-Splatting & 0.8363 & 23.28 & 0.1354 & 807.78 & 3.3600 & 1451 \\*
 & Pixel-GS & 0.8346 & 23.15 & \textbf{0.1340} & 841.11 & 3.5600 & 1321 \\*
 & GDAGS & 0.8339 & 22.69 & 0.1384 & 554.20 & 2.6700 & 829 \\*
 & Mini-Splatting & \textbf{0.8451} & 23.20 & 0.1351 & 938.93 & 3.9700 & 1207 \\*
 & ACD-GS & 0.8450 & \textbf{23.49} & 0.1452 & \textbf{223.28} & \textbf{0.9400} & 691 \\
\midrule
Museum-2 & AbsGS & 0.8798 & 25.48 & 0.1109 & 310.55 & 1.3100 & 782 \\*
 & 3DGS & 0.8970 & 25.34 & 0.0985 & 333.72 & 1.4100 & 904 \\*
 & Mip-Splatting & 0.8964 & 25.81 & 0.0976 & 477.92 & 1.9900 & 1045 \\*
 & Pixel-GS & 0.8982 & 26.58 & 0.0932 & 561.85 & 2.3800 & 1006 \\*
 & GDAGS & 0.8988 & 25.64 & 0.0951 & 299.60 & 1.2900 & 676 \\*
 & Mini-Splatting & \textbf{0.9053} & 27.50 & \textbf{0.0891} & 877.42 & 3.9700 & 1140 \\*
 & ACD-GS & 0.9002 & \textbf{27.61} & 0.1056 & \textbf{121.68} & \textbf{0.5100} & \textbf{557} \\
\midrule
NightSnow & AbsGS & 0.9033 & 30.09 & 0.2016 & 145.95 & 0.6200 & 601 \\*
 & 3DGS & 0.9095 & 30.53 & 0.1962 & 166.04 & 0.7000 & 625 \\*
 & Mip-Splatting & \textbf{0.9160} & 30.37 & \textbf{0.1765} & 251.49 & 1.0500 & 747 \\*
 & Pixel-GS & 0.9119 & 30.13 & 0.1925 & 285.68 & 1.2100 & 658 \\*
 & GDAGS & 0.9153 & \textbf{30.63} & 0.1857 & 131.93 & 0.5400 & 517 \\*
 & Mini-Splatting & 0.9080 & 29.56 & 0.1901 & 877.21 & 3.7100 & 1113 \\*
 & ACD-GS & 0.9054 & 30.22 & 0.2200 & \textbf{64.62} & \textbf{0.2700} & \textbf{465} \\
\midrule
Playroom & AbsGS & 0.8893 & 30.00 & 0.2644 & 405.12 & 1.7100 & 679 \\*
 & 3DGS & 0.8919 & 29.99 & 0.2763 & 499.96 & 2.1100 & 856 \\*
 & Mip-Splatting & 0.8906 & 29.68 & 0.2688 & 797.40 & 3.3200 & 1044 \\*
 & Pixel-GS & 0.8892 & 29.62 & 0.2763 & 959.15 & 4.0600 & 995 \\*
 & GDAGS & 0.8949 & \textbf{30.22} & \textbf{0.2634} & 444.93 & 1.9200 & 544 \\*
 & Mini-Splatting & 0.8858 & 29.59 & 0.2979 & 1080.89 & 4.5700 & 962 \\*
 & ACD-GS & \textbf{0.8960} & 30.07 & 0.2883 & \textbf{129.12} & \textbf{0.5500} & \textbf{432} \\
\midrule
Ponche & AbsGS & 0.8000 & 22.31 & 0.2376 & 254.42 & 1.0800 & 690 \\*
 & 3DGS & 0.8204 & 22.25 & 0.2464 & 205.92 & 0.8700 & 672 \\*
 & Mip-Splatting & 0.8201 & 21.80 & 0.2250 & 337.44 & 1.4000 & 842 \\*
 & Pixel-GS & 0.8123 & 21.57 & 0.2360 & 404.74 & 1.7100 & 834 \\*
 & GDAGS & 0.8148 & 22.09 & 0.2333 & 225.64 & 0.9800 & 589 \\*
 & Mini-Splatting & \textbf{0.8341} & 22.86 & \textbf{0.2067} & 848.29 & 3.5900 & 1094 \\*
 & ACD-GS & 0.8191 & \textbf{23.07} & 0.2537 & \textbf{114.83} & \textbf{0.4900} & \textbf{543} \\
\midrule
SaintAnne & AbsGS & 0.8054 & 25.09 & 0.3259 & 286.99 & 1.2100 & 704 \\*
 & 3DGS & 0.8208 & 25.60 & 0.3398 & 171.70 & 0.7300 & 814 \\*
 & Mip-Splatting & 0.8216 & 25.27 & 0.3238 & 261.31 & 1.0900 & 819 \\*
 & Pixel-GS & \textbf{0.8251} & \textbf{25.67} & 0.3082 & 757.02 & 3.2000 & 1055 \\*
 & GDAGS & 0.8232 & 25.58 & \textbf{0.3078} & 252.59 & 1.2100 & 593 \\*
 & Mini-Splatting & 0.8177 & 25.44 & 0.3163 & 893.32 & 3.7800 & 1160 \\*
 & ACD-GS & 0.8203 & 25.51 & 0.3309 & \textbf{160.94} & \textbf{0.6800} & \textbf{576} \\
\midrule
Shed & AbsGS & 0.9420 & 32.32 & 0.1593 & 349.88 & 1.4800 & 713 \\*
 & 3DGS & 0.9474 & 32.22 & 0.1657 & 312.16 & 1.3200 & 735 \\*
 & Mip-Splatting & 0.9489 & 32.32 & \textbf{0.1452} & 526.94 & 2.1900 & 912 \\*
 & Pixel-GS & \textbf{0.9507} & \textbf{32.58} & 0.1487 & 664.38 & 2.8100 & 906 \\*
 & GDAGS & 0.9501 & 32.34 & 0.1474 & 307.70 & 1.1800 & 563 \\*
 & Mini-Splatting & 0.9384 & 30.76 & 0.1547 & 983.32 & 4.1600 & 1126 \\*
 & ACD-GS & 0.9295 & 27.28 & 0.1963 & \textbf{107.75} & \textbf{0.4600} & \textbf{490} \\
\midrule
Tree-18 & AbsGS & 0.7458 & 21.03 & 0.2105 & 216.07 & 0.9100 & 751 \\*
 & 3DGS & 0.7974 & 20.75 & 0.1695 & 223.22 & 0.9400 & 838 \\*
 & Mip-Splatting & 0.7887 & 21.06 & 0.1765 & 289.15 & 1.2000 & 946 \\*
 & Pixel-GS & 0.7958 & 21.76 & 0.1668 & 327.29 & 1.3800 & 884 \\*
 & GDAGS & 0.7705 & 21.26 & 0.2049 & 192.28 & 0.8800 & 679 \\*
 & Mini-Splatting & 0.8036 & 21.98 & 0.1714 & 798.00 & 3.3700 & 1167 \\*
 & ACD-GS & \textbf{0.8052} & \textbf{22.22} & \textbf{0.1650} & \textbf{84.78} & \textbf{0.3600} & \textbf{639} \\
\midrule
Yellowhouse-12 & AbsGS & 0.5812 & 16.29 & 0.3517 & 236.54 & 1.0000 & 655 \\*
 & 3DGS & 0.6554 & 17.11 & 0.2949 & 176.08 & 0.7400 & 642 \\*
 & Mip-Splatting & 0.6414 & 16.75 & 0.3157 & 238.22 & 0.9900 & 780 \\*
 & Pixel-GS & \textbf{0.6670} & 17.61 & \textbf{0.2816} & 267.99 & 1.1300 & 709 \\*
 & GDAGS & 0.6371 & 16.63 & 0.3081 & 180.94 & 0.8500 & 575 \\*
 & Mini-Splatting & 0.6394 & 16.27 & 0.3155 & 749.61 & 4.1600 & 970 \\*
 & ACD-GS & 0.6608 & \textbf{18.09} & 0.2975 & \textbf{95.58} & \textbf{0.4000} & \textbf{537} \\
\midrule
Mean & AbsGS & 0.8171 & 24.86 & 0.2433 & 400.42 & 1.6900 & 795 \\*
 & 3DGS & 0.8448 & 25.34 & 0.2265 & 367.25 & 1.5500 & 842 \\*
 & Mip-Splatting & 0.8418 & 25.15 & 0.2211 & 563.20 & 2.3400 & 1006 \\*
 & Pixel-GS & 0.8440 & \textbf{25.63} & \textbf{0.2208} & 694.85 & 2.9400 & 977 \\*
 & GDAGS & 0.8384 & 25.22 & 0.2250 & 360.60 & 1.5800 & 628 \\*
 & Mini-Splatting & 0.8349 & 24.94 & 0.2392 & 959.15 & 3.3700 & 1125 \\*
 & ACD-GS & \textbf{0.8451} & 25.59 & 0.2363 & \textbf{151.85} & \textbf{0.6400} & \textbf{545} \\
\midrule
\end{longtable}
\endgroup

%% file: tables/appendix_combined_05_v8.tex
\begingroup\small
\begin{longtable}{@{}llcccccc@{}}
\caption{Per-scene quality and efficiency on Mip-NeRF 360.}\label{tab:mipnerf360_combined}\\
\toprule
Scene & Method & SSIM$\uparrow$ & PSNR$\uparrow$ & LPIPS$\downarrow$ & Size(MB)$\downarrow$ & Points(M)$\downarrow$ & Time(s)$\downarrow$ \\
\midrule
\endfirsthead
\multicolumn{8}{l}{\tablename~\thetable{} (continued)}\\
\toprule
Scene & Method & SSIM$\uparrow$ & PSNR$\uparrow$ & LPIPS$\downarrow$ & Size(MB)$\downarrow$ & Points(M)$\downarrow$ & Time(s)$\downarrow$ \\
\midrule
\endhead
\midrule
\multicolumn{8}{r}{Continued on next page}\\
\endfoot
\bottomrule
\endlastfoot
Bicycle & AbsGS & 0.7817 & 25.26 & 0.1720 & 1411.36 & 5.9700 & 1084 \\*
 & 3DGS & 0.7666 & 25.28 & 0.2081 & 1154.55 & 4.8800 & 1084 \\*
 & Mip-Splatting & \textbf{0.7924} & \textbf{25.55} & \textbf{0.1669} & 1977.66 & 8.2300 & 1674 \\*
 & Pixel-GS & 0.7817 & 25.26 & 0.1720 & 2142.21 & 9.0600 & 1639 \\*
 & GDAGS & 0.7899 & 25.42 & 0.1696 & 1437.29 & 6.0800 & 807 \\*
 & Mini-Splatting & 0.7165 & 23.57 & 0.2508 & 1425.72 & 6.0300 & 1031 \\*
 & ACD-GS & 0.7861 & 25.36 & 0.1799 & \textbf{543.99} & \textbf{2.3000} & \textbf{597} \\
\midrule
Bonsai & AbsGS & 0.9575 & 32.67 & 0.0847 & 198.66 & 0.8400 & 360 \\*
 & 3DGS & 0.9434 & 31.54 & 0.0964 & 253.60 & 1.0700 & 360 \\*
 & Mip-Splatting & \textbf{0.9588} & \textbf{32.95} & \textbf{0.0819} & 385.98 & 1.6100 & 485 \\*
 & Pixel-GS & 0.9575 & 32.67 & 0.0847 & 481.79 & 2.0400 & 499 \\*
 & GDAGS & 0.9552 & 32.22 & 0.0901 & 184.95 & 0.7800 & 261 \\*
 & Mini-Splatting & 0.9353 & 29.05 & 0.1055 & 882.88 & 3.7300 & 715 \\*
 & ACD-GS & 0.9455 & 31.07 & 0.1143 & \textbf{76.09} & \textbf{0.3200} & \textbf{249} \\
\midrule
Counter & AbsGS & 0.9216 & 29.35 & 0.1073 & 190.66 & 0.8100 & 411 \\*
 & 3DGS & 0.9212 & 28.43 & 0.1121 & 235.90 & 1.0000 & 411 \\*
 & Mip-Splatting & \textbf{0.9249} & \textbf{29.62} & \textbf{0.1042} & 343.31 & 1.4300 & 494 \\*
 & Pixel-GS & 0.9216 & 29.35 & 0.1073 & 574.00 & 2.4300 & 613 \\*
 & GDAGS & 0.9174 & 28.95 & 0.1136 & 188.99 & 0.8000 & 283 \\*
 & Mini-Splatting & 0.8804 & 26.38 & 0.1465 & 880.19 & 3.7200 & 770 \\*
 & ACD-GS & 0.9127 & 28.83 & 0.1251 & \textbf{69.44} & \textbf{0.2900} & \textbf{254} \\
\midrule
Flowers & AbsGS & 0.6210 & 21.26 & \textbf{0.2706} & 907.98 & 3.8400 & 732 \\*
 & 3DGS & 0.6044 & 21.50 & 0.3366 & 686.30 & 2.9000 & 732 \\*
 & Mip-Splatting & \textbf{0.6404} & \textbf{21.68} & 0.2732 & 1104.81 & 4.6000 & 1044 \\*
 & Pixel-GS & 0.6210 & 21.26 & \textbf{0.2706} & 1774.43 & 7.5000 & 1433 \\*
 & GDAGS & 0.6224 & 21.34 & 0.2746 & 878.11 & 3.7100 & 622 \\*
 & Mini-Splatting & 0.5724 & 20.85 & 0.3516 & 1152.61 & 4.8700 & 967 \\*
 & ACD-GS & 0.6199 & 21.32 & 0.2811 & \textbf{379.47} & \textbf{1.6000} & \textbf{530} \\
\midrule
Garden & AbsGS & 0.8716 & 27.55 & 0.0991 & 896.76 & 3.7900 & 1003 \\*
 & 3DGS & 0.8680 & 27.45 & 0.1063 & 981.85 & 4.1500 & 1003 \\*
 & Mip-Splatting & \textbf{0.8772} & \textbf{27.66} & \textbf{0.0945} & 1499.50 & 6.2400 & 1472 \\*
 & Pixel-GS & 0.8716 & 27.55 & 0.0991 & 2075.27 & 8.7700 & 1654 \\*
 & GDAGS & 0.8715 & 27.44 & 0.1006 & 1042.35 & 4.4100 & 767 \\*
 & Mini-Splatting & 0.8300 & 25.73 & 0.1476 & 1375.65 & 5.8200 & 1136 \\*
 & ACD-GS & 0.8689 & 27.49 & 0.1056 & \textbf{461.87} & \textbf{1.9500} & \textbf{613} \\
\midrule
Kitchen & AbsGS & 0.9495 & \textbf{32.58} & 0.0652 & 245.16 & 1.0400 & 483 \\*
 & 3DGS & 0.9488 & 32.25 & 0.0667 & 363.99 & 1.5400 & 483 \\*
 & Mip-Splatting & \textbf{0.9498} & 32.26 & \textbf{0.0638} & 492.76 & 2.0500 & 589 \\*
 & Pixel-GS & 0.9495 & \textbf{32.58} & 0.0652 & 720.13 & 3.0400 & 710 \\*
 & GDAGS & 0.9483 & 32.12 & 0.0676 & 249.48 & 1.0500 & 321 \\*
 & Mini-Splatting & 0.9208 & 28.93 & 0.0969 & 864.55 & 3.6600 & 791 \\*
 & ACD-GS & 0.9359 & 30.59 & 0.0843 & \textbf{84.22} & \textbf{0.3600} & \textbf{258} \\
\midrule
Room & AbsGS & 0.9448 & 32.23 & 0.1068 & 268.59 & 1.1400 & 400 \\*
 & 3DGS & 0.9343 & 32.40 & 0.1118 & 272.45 & 1.1500 & 400 \\*
 & Mip-Splatting & \textbf{0.9464} & \textbf{32.49} & \textbf{0.1040} & 452.25 & 1.8800 & 509 \\*
 & Pixel-GS & 0.9448 & 32.23 & 0.1068 & 581.21 & 2.4600 & 542 \\*
 & GDAGS & 0.9431 & 31.97 & 0.1097 & 262.01 & 1.1100 & 275 \\*
 & Mini-Splatting & 0.9238 & 29.64 & 0.1347 & 966.48 & 4.0900 & 730 \\*
 & ACD-GS & 0.9385 & 31.87 & 0.1230 & \textbf{62.42} & \textbf{0.2600} & \textbf{227} \\
\midrule
Stump & AbsGS & 0.7768 & 26.66 & 0.1967 & 1106.18 & 4.6800 & 869 \\*
 & 3DGS & 0.7714 & 26.66 & 0.2163 & 1026.20 & 4.3400 & 869 \\*
 & Mip-Splatting & 0.7902 & 26.89 & 0.1883 & 1440.26 & 5.9900 & 1218 \\*
 & Pixel-GS & 0.7768 & 26.66 & 0.1967 & 1618.47 & 6.8400 & 1262 \\*
 & GDAGS & \textbf{0.7943} & \textbf{26.96} & \textbf{0.1874} & 949.72 & 4.0200 & 613 \\*
 & Mini-Splatting & 0.7601 & 26.03 & 0.2282 & 1274.58 & 5.3900 & 949 \\*
 & ACD-GS & 0.7830 & 26.83 & 0.2051 & \textbf{318.65} & \textbf{1.3500} & \textbf{455} \\
\midrule
Treehill & AbsGS & 0.6139 & 21.96 & 0.2795 & 1221.96 & 5.1700 & 807 \\*
 & 3DGS & 0.6335 & \textbf{22.49} & 0.3251 & 770.88 & 3.2600 & 807 \\*
 & Mip-Splatting & \textbf{0.6387} & 22.34 & \textbf{0.2744} & 1223.14 & 5.0900 & 1118 \\*
 & Pixel-GS & 0.6139 & 21.96 & 0.2795 & 1868.62 & 7.9000 & 1396 \\*
 & GDAGS & 0.6246 & 22.06 & 0.2786 & 1171.14 & 4.9500 & 700 \\*
 & Mini-Splatting & 0.6244 & 22.05 & 0.3254 & 1154.33 & 4.8800 & 944 \\*
 & ACD-GS & 0.6253 & 21.79 & 0.2936 & \textbf{472.37} & \textbf{2.0000} & \textbf{552} \\
\midrule
Mean & AbsGS & 0.8265 & 27.72 & 0.1535 & 716.37 & 3.0300 & 683 \\*
 & 3DGS & 0.8213 & 27.56 & 0.1755 & 638.41 & 2.7000 & 683 \\*
 & Mip-Splatting & \textbf{0.8354} & \textbf{27.94} & \textbf{0.1501} & 991.07 & 4.1200 & 955 \\*
 & Pixel-GS & 0.8265 & 27.72 & 0.1535 & 1315.13 & 5.5600 & 1083 \\*
 & GDAGS & 0.8296 & 27.61 & 0.1546 & 764.32 & 2.9900 & 516 \\*
 & Mini-Splatting & 0.7960 & 25.80 & 0.1986 & 1108.55 & 4.6900 & 892 \\*
 & ACD-GS & 0.8240 & 27.24 & 0.1680 & \textbf{274.28} & \textbf{1.1600} & \textbf{415} \\
\midrule
\end{longtable}
\endgroup

%% file: tables/appendix_combined_06_v8.tex
\begingroup\small
\begin{longtable}{@{}llcccccc@{}}
\caption{Per-scene quality and efficiency on Tanks \& Temples.}\label{tab:tandt_combined}\\
\toprule
Scene & Method & SSIM$\uparrow$ & PSNR$\uparrow$ & LPIPS$\downarrow$ & Size(MB)$\downarrow$ & Points(M)$\downarrow$ & Time(s)$\downarrow$ \\
\midrule
\endfirsthead
\multicolumn{8}{l}{\tablename~\thetable{} (continued)}\\
\toprule
Scene & Method & SSIM$\uparrow$ & PSNR$\uparrow$ & LPIPS$\downarrow$ & Size(MB)$\downarrow$ & Points(M)$\downarrow$ & Time(s)$\downarrow$ \\
\midrule
\endhead
\midrule
\multicolumn{8}{r}{Continued on next page}\\
\endfoot
\bottomrule
\endlastfoot
Train & AbsGS & 0.8166 & 21.64 & 0.1942 & 230.86 & 0.9800 & 401 \\*
 & 3DGS & 0.8202 & 22.04 & 0.1963 & 258.35 & 1.0900 & 510 \\*
 & Mip-Splatting & 0.8266 & 21.87 & 0.1893 & 354.45 & 1.4700 & 611 \\*
 & Pixel-GS & \textbf{0.8279} & \textbf{22.13} & \textbf{0.1780} & 906.22 & 3.8300 & 838 \\*
 & GDAGS & 0.8129 & 21.53 & 0.2068 & 223.96 & 0.9500 & \textbf{326} \\*
 & Mini-Splatting & 0.7099 & 18.77 & 0.3234 & 937.32 & 3.9600 & 855 \\*
 & ACD-GS & 0.8103 & 21.54 & 0.2298 & \textbf{123.27} & \textbf{0.5200} & 335 \\
\midrule
Truck & AbsGS & 0.8874 & 25.71 & 0.1314 & 388.02 & 1.6400 & 464 \\*
 & 3DGS & 0.8851 & 25.47 & 0.1418 & 487.82 & 2.0600 & 593 \\*
 & Mip-Splatting & \textbf{0.8930} & \textbf{25.74} & 0.1231 & 784.70 & 3.2700 & 820 \\*
 & Pixel-GS & 0.8868 & 25.48 & \textbf{0.1203} & 1228.62 & 5.1900 & 1053 \\*
 & GDAGS & 0.8858 & 25.49 & 0.1380 & 415.07 & 1.7500 & 374 \\*
 & Mini-Splatting & 0.8573 & 23.66 & 0.1583 & 1084.63 & 4.5900 & 859 \\*
 & ACD-GS & 0.8813 & 25.21 & 0.1434 & \textbf{177.46} & \textbf{0.7500} & \textbf{331} \\
\midrule
Mean & AbsGS & 0.8520 & 23.68 & 0.1628 & 309.44 & 1.3100 & 432 \\*
 & 3DGS & 0.8527 & 23.76 & 0.1691 & 373.09 & 1.5800 & 551 \\*
 & Mip-Splatting & \textbf{0.8598} & 23.80 & 0.1562 & 569.58 & 2.3700 & 715 \\*
 & Pixel-GS & 0.8574 & \textbf{23.81} & \textbf{0.1492} & 1067.42 & 4.5100 & 945 \\*
 & GDAGS & 0.8494 & 23.51 & 0.1724 & 319.51 & 1.3500 & 350 \\*
 & Mini-Splatting & 0.7836 & 21.22 & 0.2409 & 1010.98 & 4.2700 & 857 \\*
 & ACD-GS & 0.8458 & 23.37 & 0.1866 & \textbf{150.37} & \textbf{0.6400} & \textbf{333} \\
\midrule
\end{longtable}
\endgroup

%% file: geometry_v8.tex
\section{Geometric Reconstruction}
\label{sec:geometry}

To verify whether the proposed geometry-aware densification preserves reconstructed geometry, we conduct an additional surface reconstruction evaluation. Specifically, we extract meshes from the reconstructed Gaussian representations and evaluate them using Precision, Recall, and F-score. The results are reported in Table~\ref{tab:geometry}.

\begin{table}[htbp]
  \centering
  \caption{Geometric reconstruction evaluation using Precision, Recall, and F-score.}
  \label{tab:geometry}
  \small
  \begin{tabular}{@{}lcccc@{}}
    \toprule
    Method & Precision(\%)$\uparrow$ & Recall(\%)$\uparrow$ & F-score(\%)$\uparrow$ & Points$\downarrow$ \\
    \midrule
ACD-GS & 66.34 & 91.21 & 76.81 & \textbf{642,044}\\
AbsGS & \textbf{69.58} & \textbf{93.33} & \textbf{79.72} & 1,693,016\\
3DGS & 60.52 & 79.08 & 68.57 & 1,552,777\\
GDAGS & 66.13 & 88.80 & 75.81 & 1,582,823\\
Mini-Splatting & 56.04 & 85.58 & 67.72 & 4,495,856\\
Mip-Splatting & 66.02 & 88.60 & 78.03 & 2,343,461\\
Pixel-GS & 68.16 & 86.41 & 76.21 & 2,937,890\\
    \bottomrule
  \end{tabular}
\end{table}

Our method primarily aims to obtain a more compact Gaussian representation while maintaining comparable novel-view synthesis quality, without introducing additional geometric supervision or explicit constraints on reconstruction accuracy. ACD-GS uses only approximately 38\% and 41\% of the Gaussian primitives required by AbsGS and GDAGS, respectively. Moreover, because ACD-GS tends to represent planar regions with fewer and larger Gaussians, TSDF-based surface extraction may be more susceptible to holes and local reconstruction errors. Nevertheless, ACD-GS still achieves competitive geometric reconstruction quality, indicating that curvature-aware densification preserves essential scene structures despite the substantially reduced Gaussian budget.

%% file: checklist.tex
\section*{NeurIPS Paper Checklist}

\begin{enumerate}

\item {\bf Claims}
    \item[] Question: Do the main claims made in the abstract and introduction accurately reflect the paper's contributions and scope?
    \item[] Answer: \answerYes{} 
    \item[] Justification:  The abstract and introduction reflect the paper’s contributions and scope.

\item {\bf Limitations}
    \item[] Question: Does the paper discuss the limitations of the work performed by the authors?
    \item[] Answer: \answerYes{} 
    \item[] Justification: We discuss the limitations of this work in a single section.

\item {\bf Theory assumptions and proofs}
    \item[] Question: For each theoretical result, does the paper provide the full set of assumptions and a complete (and correct) proof?
    \item[] Answer: \answerYes{} 
    \item[] Justification: The assumptions for each theoretical result are clearly stated.

    \item {\bf Experimental result reproducibility}
    \item[] Question: Does the paper fully disclose all the information needed to reproduce the main experimental results of the paper to the extent that it affects the main claims and/or conclusions of the paper (regardless of whether the code and data are provided or not)?
    \item[] Answer: \answerYes{} 
    \item[] Justification: We provide sufficient information to reproduce the main experimental results and the code will be released.

\item {\bf Open access to data and code}
    \item[] Question: Does the paper provide open access to the data and code, with sufficient instructions to faithfully reproduce the main experimental results, as described in supplemental material?
    \item[] Answer: \answerYes{} 
    \item[] Justification: We provide the implementation code of our method and model checkpoints.

\item {\bf Experimental setting/details}
    \item[] Question: Does the paper specify all the training and test details (e.g., data splits, hyperparameters, how they were chosen, type of optimizer) necessary to understand the results?
    \item[] Answer: \answerYes{}
    \item[] Justification: We specify all the training and test details.

\item {\bf Experiment statistical significance}
    \item[] Question: Does the paper report error bars suitably and correctly defined or other appropriate information about the statistical significance of the experiments?
    \item[] Answer: \answerNo{} 
    \item[] Justification:  The paper does not include hypothesis testing or statistical significance analysis, and error bars are not a focus of the experiments.

\item {\bf Experiments compute resources}
    \item[] Question: For each experiment, does the paper provide sufficient information on the computer resources (type of compute workers, memory, time of execution) needed to reproduce the experiments?
    \item[] Answer: \answerYes{} 
    \item[] Justification:  The computational resources are discussed in the Implementation section,
providing sufficient information to reproduce the experiments.
    
\item {\bf Code of ethics}
    \item[] Question: Does the research conducted in the paper conform, in every respect, with the NeurIPS Code of Ethics \url{https://neurips.cc/public/EthicsGuidelines}?
    \item[] Answer: \answerYes{} 
    \item[] Justification: The research conducted in the paper confirms the NeurIPS code of Ethics.

\item {\bf Broader impacts}
    \item[] Question: Does the paper discuss both potential positive societal impacts and negative societal impacts of the work performed?
    \item[] Answer: \answerNA{} 
    \item[] Justification: Our work mainly focuses on the improvement of algorithms and models, rather than areas directly related to social or ethical issues.
    
\item {\bf Safeguards}
    \item[] Question: Does the paper describe safeguards that have been put in place for responsible release of data or models that have a high risk for misuse (e.g., pre-trained language models, image generators, or scraped datasets)?
    \item[] Answer: \answerNA{} 
    \item[] Justification: The paper poses no such risks.

\item {\bf Licenses for existing assets}
    \item[] Question: Are the creators or original owners of assets (e.g., code, data, models), used in the paper, properly credited and are the license and terms of use explicitly mentioned and properly respected?
    \item[] Answer: \answerYes{} 
    \item[] Justification: All resources used in our paper, including code, data, and models, strictly
adhere to appropriate licenses and terms of use, and have been appropriately authored by
the original owner

\item {\bf New assets}
    \item[] Question: Are new assets introduced in the paper well documented and is the documentation provided alongside the assets?
    \item[] Answer: \answerYes{} 
    \item[] Justification: The code is provided and includes accompanying documentation.

\item {\bf Crowdsourcing and research with human subjects}
    \item[] Question: For crowdsourcing experiments and research with human subjects, does the paper include the full text of instructions given to participants and screenshots, if applicable, as well as details about compensation (if any)? 
    \item[] Answer: \answerNA{} 
    \item[] Justification:  This work does not involve crowdsourcing nor research with human subjects.

\item {\bf Institutional review board (IRB) approvals or equivalent for research with human subjects}
    \item[] Question: Does the paper describe potential risks incurred by study participants, whether such risks were disclosed to the subjects, and whether Institutional Review Board (IRB) approvals (or an equivalent approval/review based on the requirements of your country or institution) were obtained?
    \item[] Answer: \answerNA{} 
    \item[] Justification: the paper does not involve crowdsourcing nor research with human subjects.

\item {\bf Declaration of LLM usage}
    \item[] Question: Does the paper describe the usage of LLMs if it is an important, original, or non-standard component of the core methods in this research? Note that if the LLM is used only for writing, editing, or formatting purposes and does \emph{not} impact the core methodology, scientific rigor, or originality of the research, declaration is not required.
    \item[] Answer:\answerNA{} 
    \item[] Justification: We use LLM for writing, editing, and formatting purposes. Coding Agents are only used to manage training tasks.

\end{enumerate}